\documentclass[10pt,twocolumn,letterpaper]{article}

\usepackage[pagenumbers]{cvpr} 

\usepackage{graphicx}
\usepackage{amsmath}
\usepackage{amssymb}
\usepackage{booktabs}

\usepackage[pagebackref,breaklinks,colorlinks]{hyperref}

\usepackage[capitalize]{cleveref}
\crefname{section}{Sec.}{Secs.}
\Crefname{section}{Section}{Sections}
\Crefname{table}{Table}{Tables}
\crefname{table}{Tab.}{Tabs.}

\def\cvprPaperID{773} 
\def\confName{CVPR}
\def\confYear{2023}

\usepackage{multirow}
\usepackage{makecell}
\usepackage{booktabs} 
\usepackage{pifont}
\newcommand{\rankfirst}[1]{\textbf{#1}}
\newcommand{\ranksecond}[1]{\underline{#1}}
\newcommand{\mj}{$\mathcal{J}$}
\newcommand{\mf}{$\mathcal{F}$}
\newcommand{\mjf}{$\mathcal{J}\&\mathcal{F}$}

\newcommand{\mjs}{$\mathcal{J}_s$}
\newcommand{\mfs}{$\mathcal{F}_s$}
\newcommand{\mju}{$\mathcal{J}_u$}
\newcommand{\mfu}{$\mathcal{F}_u$}
\newcommand{\mg}{$\mathcal{G}$}

\usepackage{bm}
\def\pt{p_\textrm{t}}
\def\at{\alpha_\textrm{t}}
\newcommand{\eqnnm}[2]{\begin{equation}\label{eq:#1}#2\end{equation}\ignorespaces}
\usepackage{bbm,nicefrac}
\usepackage{colortbl}

\begin{document}

\title{Universal Instance Perception as Object Discovery and Retrieval}

\author{Bin Yan$^{1}$\thanks{This work was performed while Bin Yan worked as an intern at
ByteDance. Email: \href{mailto:yan_bin@mail.dlut.edu.cn}{\color{black}{yan\_bin@mail.dlut.edu.cn}.} $^\dagger$ Corresponding authors: \href{mailto:jiangyi.enjoy@bytedance.com}{\color{black}{jiangyi.enjoy@bytedance.com}}, \href{mailto:wdice@dlut.edu.cn}{\color{black}{wdice@dlut.edu.cn}}.},
Yi Jiang$^{2, \dagger}$,
Jiannan Wu$^{3}$, 
Dong Wang$^{1, \dagger}$, \\
Ping Luo$^{3}$,
Zehuan Yuan$^{2}$,
Huchuan Lu$^{1,4}$\\
$^{1}$ School of Information and Communication Engineering, Dalian University of
Technology, China \\
$^{2}$ ByteDance $^{3}$ The University of Hong Kong $^{4}$ Peng Cheng Laboratory
}
%
%
\maketitle

\begin{abstract}
All instance perception tasks aim at finding certain objects specified by some queries such as category names, language expressions, and target annotations, but this complete field has been split into multiple independent sub-tasks.
In this work, we present a \textbf{un}iversal \textbf{in}stance perception model of the \textbf{next} generation, termed \textbf{UNINEXT}.
UNINEXT reformulates diverse instance perception tasks into a unified object discovery and retrieval paradigm and can flexibly perceive different types of objects by simply changing the input prompts. This unified formulation brings the following benefits: (1) enormous data from different tasks and label vocabularies can be exploited for jointly training general instance-level representations, which is especially beneficial for tasks lacking in training data. (2)  the unified model is parameter-efficient and can save redundant computation when handling multiple tasks simultaneously.  
UNINEXT shows superior performance on 20 challenging benchmarks from 10 instance-level tasks including classical image-level tasks (object detection and instance segmentation), vision-and-language tasks (referring expression comprehension and segmentation), and six video-level object tracking tasks. 
Code is available at \href{https://github.com/MasterBin-IIAU/UNINEXT}{https://github.com/MasterBin-IIAU/UNINEXT}.

\end{abstract}

\section{Introduction}
\label{sec:intro}

\begin{figure}[!t]
  \begin{center}
\includegraphics[width=1.0\linewidth]{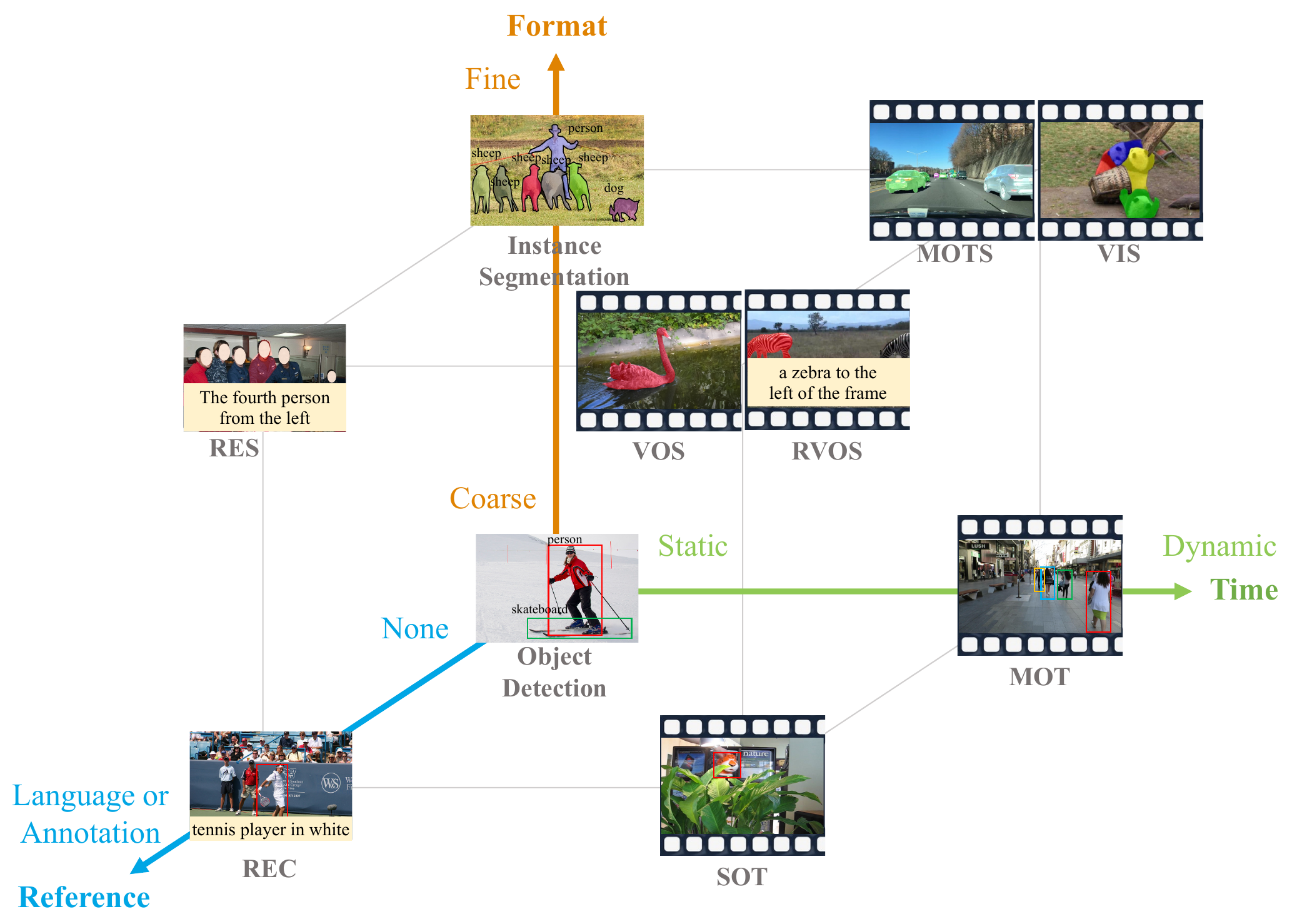}
  \end{center}
  \vspace{-5mm}
  \caption{Task distribution on the Format-Time-Reference space. Better view on screen with zoom-in.} 
  \label{fig-tasks}
\vspace{-3mm}
\end{figure}

Object-centric understanding is one of the most essential and challenging problems in computer vision. Over the years, the diversity of this field increases substantially. In this work, we mainly discuss 10 sub-tasks, distributed on the vertices of the cube shown in Figure~\ref{fig-tasks}. As the most fundamental tasks, object detection~\cite{DPM,FasterRCNN,FPN,YOLO,FCOS,CascadeRCNN, DETR} and instance segmentation~\cite{MaskRCNN,SOLO,PANet,YOLACT,CondInst} require finding all objects of specific categories by boxes and masks respectively. Extending inputs from static images to dynamic videos, Multiple Object Tracking (MOT)~\cite{MOT17,Tracktor,CenterTrack,FairMOT}, Multi-Object Tracking and Segmentation (MOTS)~\cite{MOTS,PointTrackV2,PCAN}, and Video Instance Segmentation (VIS)~\cite{VIS, VISTR, SeqFormer,IFC} require finding all object trajectories of specific categories in videos. Except for \textbf{category names}, some tasks provide other reference information. For example, Referring Expression Comprehension (REC)~\cite{RefCOCO&plus,FAOA,SeqTR}, Referring Expression Segmentation (RES)~\cite{RefCOCO&plus,CMSA,LAVT}, and Referring Video Object Segmentation (R-VOS)~\cite{URVOS,MTTR,ReferFormer} aim at finding objects matched with the given \textbf{language expressions} like ``The fourth person from the left''. Besides, Single Object Tracking (SOT)~\cite{OTB2015,SiameseFC,SiameseRPN} and Video Object Segmentation (VOS)~\cite{YoutubeVOS,STM,STCN} take the \textbf{target annotations} (boxes or masks) given in the first frame as the reference, requiring to predict the trajectories of the tracked objects in the subsequent frames. Since all the above tasks aim to perceive instances of certain properties, we refer to them collectively as \textit{instance perception}.


Although bringing convenience to specific applications, such diverse task definitions split the whole field into fragmented pieces. As the result, most current instance perception methods are developed for only a single or a part of sub-tasks and trained on data from specific domains. Such fragmented design philosophy brings the following drawbacks: (1) Independent designs hinder models from learning and sharing generic knowledge between different tasks and domains, causing redundant parameters.  (2) The possibility of mutual collaboration between different tasks is overlooked. For example, object detection data enables models to recognize common objects, which can naturally improve the performance of REC and RES. (3) Restricted by fixed-size classifiers, traditional object detectors are hard to jointly train on multiple datasets with different label vocabularies~\cite{COCO,LVIS,Objects365} and to dynamically change object categories to detect during inference~\cite{COCO,VIS,OVIS,BDD100K,MOT17,TAO}. Since \textit{essentially all instance perception tasks aim at finding certain objects according to some queries}, it leads to a natural question: could we design a unified model to solve all mainstream instance perception tasks once and for all?

To answer this question, we propose UNINEXT, a universal instance perception model of the next generation. We first reorganize 10 instance perception tasks into three types according to the different input prompts: (1) \textbf{category names as prompts} (Object Detection, Instance Segmentation, VIS, MOT, MOTS). (2) \textbf{language expressions as prompts} (REC, RES, R-VOS). (3) \textbf{reference annotations as prompts} (SOT, VOS). Then we propose a unified prompt-guided object discovery and retrieval formulation to solve all the above tasks. Specifically, \textbf{UNINEXT first discovers $\boldsymbol{N}$ object proposals under the guidance of the prompts, then retrieves the final instances from the proposals according to the instance-prompt matching scores}. Based on this new formulation, UNINEXT can flexibly perceive different instances by simply changing the input prompts. To deal with different prompt modalities, we adopt a prompt generation module, which consists of a reference text encoder and a reference visual encoder. Then an early fusion module is used to enhance the raw visual features of the current image and the prompt embeddings. This operation enables deep information exchange and provides highly discriminative representations for the later instance prediction step. Considering the flexible query-to-instance fashion, we choose a Transformer-based object detector~\cite{DeformableDETR} as the instance decoder. Specifically, the decoder first generates $N$ instance proposals, then the prompt is used to retrieve matched objects from these proposals. This flexible retrieval mechanism overcomes the disadvantages of traditional fixed-size classifiers and enables joint training on data from different tasks and domains.

With the unified model architecture, UNINEXT can learn strong generic representations on massive data from various tasks and solve 10 instance-level perception tasks using a single model with the same model parameters. Extensive experiments demonstrate that UNINEXT achieves superior performance on 20 challenging benchmarks. The contributions of our work can be summarized as follows. 
\begin{itemize}
	\vspace{-1mm}
	\item{We propose a unified prompt-guided formulation for universal instance perception, reuniting previously fragmented instance-level sub-tasks into a whole.}
	\vspace{-1mm}
	\item{
	Benefiting from the flexible object discovery and retrieval paradigm, UNINEXT can train on different tasks and domains, in no need of task-specific heads.
	}
	\vspace{-1mm}
	\item{UNINEXT achieves superior performance on 20 challenging benchmarks from 10 instance perception tasks using a single model with the same model parameters.}
\end{itemize}

\section{Related Work}
\textbf{Instance Perception.} The goals and typical methods of 10 instance perception tasks are introduced as follows. 

\textit{Retrieval by Category Names}. Object detection and instance segmentation aim at finding all objects of specific classes on the images in the format of boxes or masks. Early object detectors can be mainly divided into two-stage methods~\cite{FasterRCNN,CascadeRCNN,HTC} and one-stage methods~\cite{YOLO,RetinaNet,FCOS,ATSS,YOLOX} according to whether to use RoI-level operations~\cite{FastRCNN,MaskRCNN}. Recently, Transformer-based detectors~\cite{DETR,DeformableDETR,DN-DETR} have drawn great attention for their conceptually simple and flexible frameworks. Besides, instance segmentation approaches can also be divided into detector-based~\cite{MaskRCNN,CascadeRCNN,HTC,PointRend,CondInst} and detector-free~\cite{SOLO,Mask2Former} fashions according to whether box-level detectors are needed. Object detection and instance segmentation play critical roles and are foundations for all other instance perception tasks. 
For example, MOT, MOTS, and VIS extend image-level detection and segmentation to videos, requiring finding all object trajectories of specific classes in videos. Mainstream algorithms~\cite{DeepSORT,JDE,QDTrack,bytetrack,PCAN,Unicorn} of MOT and MOTS follow an online "detection-then-association" paradigm. However, due to the intrinsic difference in benchmarks of MOTS~\cite{MOTS,BDD100K} (high-resolution long videos) and VIS~\cite{VIS} (low-resolution short videos), most recent VIS methods~\cite{VISTR,IFC,ProposeReduce,SeqFormer} adopt an offline fashion. This strategy performs well on relatively simple VIS2019~\cite{VIS}, but the performance drops drastically on challenging OVIS~\cite{OVIS} benchmark. Recently, IDOL~\cite{IDOL} bridges the performance gap between online fashion and its offline counterparts by discriminative instance embeddings, showing the potential of the online paradigm in unifying MOT, MOTS, and VIS. 

\textit{Retrieval by Language Expressions}. REC, RES, and R-VOS aim at finding one specific target referred by a language expression using boxes or masks on the given images or videos. Similar to object detection, REC methods can be categorized into three paradigms: two-stage~\cite{CM-Att-Erase,DGA,RvG-Tree,NMTree}, one-stage~\cite{FAOA,RCCF,MCN,RESC}, and Transformer-based~\cite{TransVG,TRAR,MDETR} ones. Different from REC, RES approaches~\cite{CMSA,STEP,BRINet,EFN,CGAN,LTS,VLT} focus more on designing diverse attention mechanisms to achieve vision-language alignment. Recently, SeqTR~\cite{SeqTR} unifies REC and RES as a point prediction problem and obtains promising results. Finally, R-VOS can be seen as a natural extension of RES from images to videos. Current state-of-the-art methods~\cite{MTTR,ReferFormer} are Transformer-based and process the whole video in an offline fashion. However, the offline paradigm hinders the applications in the real world such as long videos and ongoing videos (e.g. autonomous driving).

\textit{Retrieval by Reference Annotations}. SOT and VOS first specify tracked objects on the first frame of a video using boxes or masks, then require algorithms to predict the trajectories of the tracked objects in boxes or masks respectively. The core problems of these two tasks include (1) How to extract informative target features? (2) How to fuse the target information with representations of the current frame? For the first question, most SOT methods~\cite{SiameseFC,SiameseRPN,SiamRPNplusplus,TransT,STARK} encode target information by passing a template to a siamese backbone. While VOS approaches~\cite{STM,CFBI,STCN} usually pass multiple previous frames together with corresponding mask results to a memory encoder for extracting fine-grained target information. For the second question, correlations are widely adopted by early SOT algorithms~\cite{SiameseFC,SiameseRPN,AlphaRefine}. However, these simple linear operations may cause serious information loss. To alleviate this problem, later works~\cite{TransT,STARK,MixFormer,OSTrack} resort to Transformer for more discriminative representations. Besides, feature fusion in VOS is almost dominated by space-time memory networks~\cite{STM,CFBI,STCN,XMem}.

\textbf{Unified Vision Models.} 
Recently, unified vision models~\cite{MuST,INTERN,OFA-Ali,Gato,MaskRCNN,Mask2Former,Unicorn,Pix2SeqV2,GLIP,Uni-Perceiver,Unified-IO} have drawn great attention and achieved significant progress due to their strong generalizability and flexibility. Unified vision models attempt to solve multiple vision or multi-modal tasks by a single model. Existing works can be categorized into unified learning paradigms and unified model architectures.

\textit{Unified Learning Paradigms.} These works~\cite{MuST,INTERN,OFA-Ali,Gato,Pathways,Uni-Perceiver,Unified-IO} usually present a universal learning paradigm for covering as many tasks and modalities as possible. For example, MuST~\cite{MuST} presents a multi-task self-training approach for 6 vision tasks. INTERN~\cite{INTERN} introduces a continuous learning scheme, showing strong generalization ability on 26 popular benchmarks. Unified-IO~\cite{Unified-IO} and OFA~\cite{OFA-Ali} proposes a unified sequence-to-sequence framework that can handle a variety of vision, language, and multi-modal tasks. Although these works can perform many tasks, the commonality and inner relationship among different tasks are less explored and exploited.


\textit{Unified Model Architectures}. These works~\cite{MaskRCNN,Mask2Former,GLIP,Pix2SeqV2,Unicorn} usually design a unified formulation or model architecture for a group of closely related tasks. For example, Mask R-CNN~\cite{MaskRCNN} proposes a unified network to perform object detection and instance segmentation simultaneously. Mask2Former~\cite{Mask2Former} presents a universal architecture capable of handling panoptic, instance, and semantic segmentation. Pix2SeqV2~\cite{Pix2SeqV2} designs a unified pixel-to-sequence interface for four vision tasks, namely object detection, instance segmentation, keypoint detection, and image captioning. GLIP~\cite{GLIP} cleverly reformulates object detection as phrase grounding by replacing classical classification with word-region alignment. This new formulation allows joint training on both detection and grounding data, showing strong transferability to various object-level recognition tasks. However, GLIP~\cite{GLIP} supports neither prompts in other modalities such as images \& annotations nor video-level tracking tasks. In terms of object tracking, Unicorn~\cite{Unicorn} proposes a unified solution for SOT, VOS, MOT, and MOTS, achieving superior performance on 8 benchmarks with the same model weights. However, it is still difficult for Unicorn to handle diverse label vocabularies~\cite{COCO,MOT17,BDD100K,TAO,VIS,OVIS} during training and inference. In this work, we propose a universal prompt-guided architecture for 10 instance perception tasks, conquering the drawbacks of GLIP~\cite{GLIP} and Unicorn~\cite{Unicorn} simultaneously.

\section{Approach}
Before introducing detailed methods, we first categorize existing instance perception tasks into three classes. 
\begin{itemize}
    \vspace{-1mm}
    \item{
    Object detection, instance segmentation, MOT, MOTS, and VIS take category names as prompts to find all instances of specific classes.} 
    \vspace{-1mm}
    \item{
    REC, RES, and R-VOS exploit an expression as the prompt to localize a certain target.} 
    \vspace{-1mm}
    \item{
    SOT and VOS use the annotation given in the first frame as the prompt for predicting the trajectories of the tracked target.}
\end{itemize}

\begin{figure*}[!t]
  \begin{center}
\includegraphics[width=1.0\linewidth]{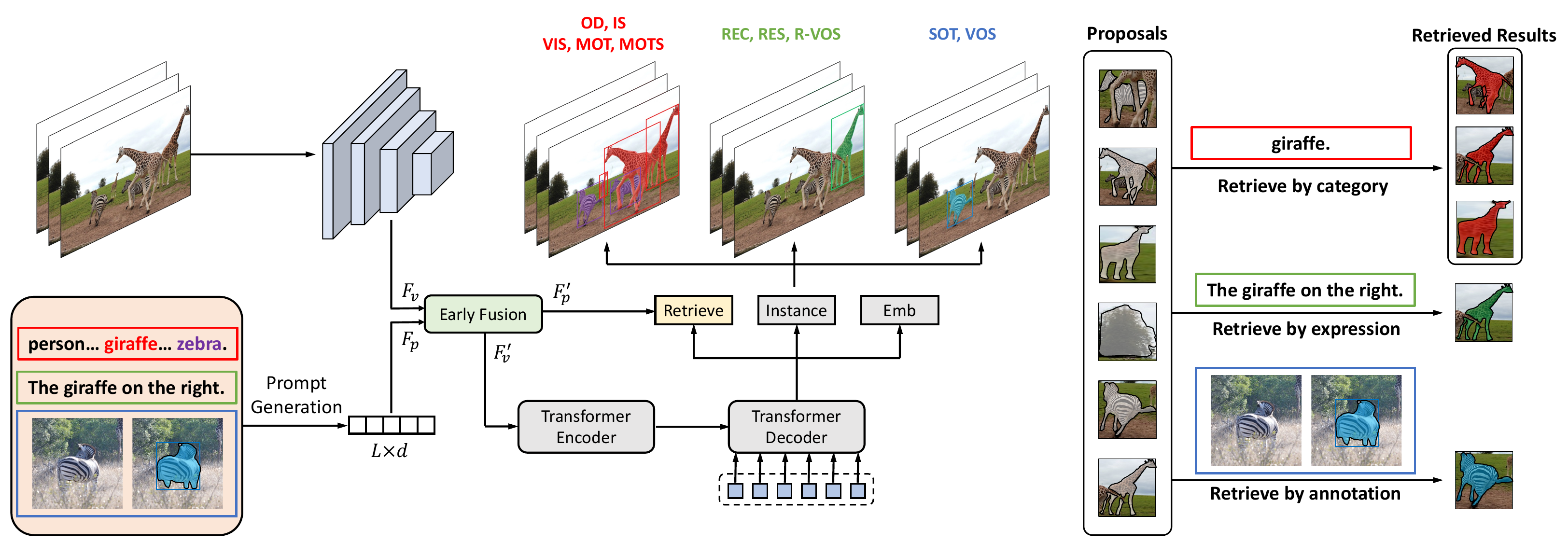}
  \end{center}
  \vspace{-5mm}
  \caption{Framework of UNINEXT. The whole pipeline is shown on the left side. The schematic diagram of object retrieval is shown on the right side. The instance head predicts both boxes and masks of the objects. Better view in color on screen.} 
  \label{fig-framework}
\vspace{-3mm}
\end{figure*}

Essentially, all the above tasks aim to find objects specified by some prompts. This commonality motivates us to reformulate all instance perception tasks into a prompt-guided object discovery and retrieval problem and solve it by a unified model architecture and learning paradigm. As demonstrated in Figure~\ref{fig-framework}, UNINEXT consists of three main components: (1) prompt generation (2) image-prompt feature fusion (3) object discovery and retrieval. 

\subsection{Prompt Generation}
First, a prompt generation module is adopted to transform the original diverse prompt inputs into a unified form. According to different modalities, we introduce the corresponding strategies in the next two paragraphs respectively. 

To deal with language-related prompts, a language encoder~\cite{BERT} $\mathrm{Enc}_\mathrm{L}$ is adopted. To be specific, for category-guided tasks, we concatenate class names that appeared in the current dataset~\cite{COCO,VIS,OVIS,BDD100K} as the language expression. Take COCO~\cite{COCO} as an example, the expression can be written as ``person. bicycle. ... . toothbrush". Then for both category-guided and expression-guided tasks, the language expression is passed into $\mathrm{Enc}_\mathrm{L}$, getting a prompt embedding ${{F}_{p}} \in\mathbb{R}^{L\times{d}}$ with a sequence length of $L$. 

For the annotation-guided tasks, to extract fine-grained visual features and fully exploit the target annotations, an additional reference visual encoder $\mathrm{Enc}^\mathrm{ref}_\mathrm{V}$ is introduced. Specifically, first a template with $2^2$ times the target box area is cropped centered on the target location on the reference frame. Then the template is resized to a fixed size of $256\times256$. To introduce more precise target information, an extra channel named the target prior is concatenated to the template image, forming a 4-channel input. In more detail, the value of the target prior is $1$ on the target region otherwise $0$. Then the template image together with the target prior is passed to the reference visual encoder $\mathrm{Enc}^\mathrm{ref}_\mathrm{V}$, obtaining a hierarchical feature pyramid $\{\mathrm{C}_3,\mathrm{C}_4,\mathrm{C}_5,\mathrm{C}_6\}$. The corresponding spatial sizes are $32\times32$, $16\times16$, $8\times8$, and $4\times4$. To keep fine target information and get the prompt embedding in the same format as other tasks, a merging module is applied. Namely, all levels of features are first upsampled to $32\times32$ then added, and flattened as the final prompt embedding ${{F}_{p}}\in\mathbb{R}^{1024\times{d}}$.

The prompt generation process can be formulated as 

\begin{align}
{{F}_{p}}=\left\{
\begin{aligned}
& \text{Enc}^{\text{ref}}_{\text{L}}(\text{expression}) & & \text{expression-guided} \\
& \text{Enc}^{\text{ref}}_{\text{L}}(\text{concat}(\text{categories})) & & \text{category-guided} \\
& \text{merge}(\text{Enc}^{\text{ref}}_{\text{V}}(\text{[template, prior]}) & & \text{annotation-guided} \nonumber\\
\end{aligned}
\right.\label{eq:prompt}
\end{align}

\subsection{Image-Prompt Feature Fusion}
In parallel with the prompt generation, the whole current image is passed through another visual encoder $\mathrm{Enc}_\mathrm{V}$, obtaining hierarchical visual features ${F}_{v}$. To enhance the original prompt embedding by the image contexts and to make the original visual features prompt-aware, an early fusion module is adopted. To be specific, first a bi-directional cross-attention module (Bi-XAtt) is used to retrieve information from different inputs, and then the retrieved representations are added to the original features. This process can be formulated as  
\begin{align}
\begin{aligned}
&F_\text{p2v}, F_\text{v2p} = \text{Bi-XAtt}(F_v, F_p)\\
&F^{\prime}_v = F_v + F_\text{p2v};\ 
F^{\prime}_{p} = F_{p} + F_\text{v2p}
\end{aligned}
\end{align}
Different from GLIP~\cite{GLIP}, which adopts 6 vision-language fusion layers and 6 additional BERT layers for feature enhancement, our early fusion module is much more efficient.

\subsection{Object Discovery and Retrieval}

With discriminative visual and prompt representations, the next crucial step is to transform input features into instances for various perception tasks. UNINEXT adopts the encoder-decoder architecture proposed by Deformable DETR~\cite{DeformableDETR} for its flexible query-to-instance fashion. We introduce the detailed architectures as follows.

The Transformer encoder takes hierarchical prompt-aware visual features as the inputs. With the help of efficient Multi-scale Deformable Self-Attention~\cite{DeformableDETR}, target information from different scales can be fully exchanged, bringing stronger instance features for the subsequent instance decoding. Besides, as performed in two-stage Deformable DETR~\cite{DeformableDETR}, an auxiliary prediction head is appended at the end of the encoder, generating $N$ initial reference points with the highest scores as the inputs of the decoder.

The Transformer decoder takes the enhanced multi-scale features, $N$ reference points from the encoder, as well as $N$ object queries as the inputs. As shown in previous works ~\cite{ReferFormer,Trackformer,VISTR,DINO}, object queries play a critical role in instance perception tasks. In this work, we attempt two query generation strategies: (1) static queries which do not change with images or prompts. (2) dynamic queries conditioned on the prompts. The first strategy can be easily implemented with \texttt{nn.Embedding(N,d)}. The second one can be performed by first pooling the enhanced prompt features ${F}^{\prime}_{v}$ along the sequence dimension, getting a global representation, then repeating it by $N$ times. The above two methods are compared in Sec~\ref{sec:ablation} and we find that static queries usually perform better than dynamic queries. The potential reason could be that static queries contain richer information and possess better training stability than dynamic queries. With the help of the deformable attention, the object queries can efficiently retrieve prompt-aware visual features and learn strong instance embedding ${F}_\mathrm{ins}\in\mathbb{R}^{N\times{d}}$.

At the end of the decoder, a group of prediction heads is exploited to obtain the final instance predictions. Specifically, an instance head produces both boxes and masks of the targets. Besides, an embedding head~\cite{IDOL} is introduced for associating the current detected results with previous trajectories in MOT, MOTS, and VIS. Until now, we have mined $N$ potential instance proposals, which are represented with gray masks in Figure~\ref{fig-framework}. However, not all proposals are what the prompts really refer to. Therefore, we need to further retrieve truly matched objects from these proposals according to the prompt embeddings as demonstrated in the right half of Figure~\ref{fig-framework}. Specifically, given the prompt embeddings $F^{\prime}_{p}$ after early fusion, for category-guided tasks, we take the embedding of each category name as a weight matrix $W\in\mathbb{R}^{1\times{d}}$. Besides, for expression-guided and annotation-guided tasks, the weight matrix $W$ is obtained by aggregating the prompt embedding $F^{\prime}_{p}$ using global average pooling (GAP) along the sequence dimension. 

\begin{align}
{W}=\left\{
\begin{aligned}
& F^{\prime}_{p}[i], \ i\in\{0,1,...,C-1\} &  & \text{category} \\
& \frac{1}{L}\sum_{i=0}^{L}F_p^\prime(i,j) & & \text{expression/annotation}\nonumber \\
\end{aligned}
\right.
\end{align}
Finally, the instance-prompt matching scores $S$ can be computed as the matrix multiplication of the target features and the transposed weight matrix. $S={F}_\mathrm{ins}W^\top$. Following previous work~\cite{GLIP}, the matching scores can be supervised by Focal Loss~\cite{RetinaNet}. Different from previous fixed-size classifiers~\cite{DeformableDETR}, the proposed retrieval head selects objects by the prompt-instance matching mechanism. This flexible design enables UNINEXT to jointly train on enormous datasets with diverse label vocabularies from different tasks, learning universal instance representations.


\subsection{Training and Inference}
\label{sub-sec-train-infer}
\textbf{Training.} The whole training process consists of three consecutive stages: (1) general perception pretraining (2) image-level joint training (3) video-level joint training. 

In the first stage, we pretrain UNINEXT on the large-scale object detection dataset Objects365~\cite{Objects365} for learning universal knowledge about objects. Since Objects365 does not have mask annotations, we introduce two auxiliary losses proposed by BoxInst~\cite{BoxInst} for training the mask branch. The loss function can be formulated as 
\begin{align}
\mathcal{L}_\mathrm{stage1} = \mathcal{L}_\mathrm{retrieve} + \mathcal{L}_\mathrm{box} + \mathcal{L}^\mathrm{boxinst}_\mathrm{mask}
\end{align}
Then based on the pretrained weights of the first stage, we finetune UNINEXT jointly on image datasets, namely COCO~\cite{COCO} and the mixed dataset of RefCOCO~\cite{RefCOCO&plus}, RefCOCO+~\cite{RefCOCO&plus}, and RefCOCOg~\cite{RefCOCOg-umd}. With manually labeled mask annotations, the traditional loss functions like Dice Loss~\cite{DiceLoss} and Focal Loss~\cite{RetinaNet} can be used for the mask learning. After this step, UNINEXT can achieve superior performance on object detection, instance segmentation, REC, and RES. 
\begin{align}
\mathcal{L}_\mathrm{stage2} = \mathcal{L}_\mathrm{retrieve} + \mathcal{L}_\mathrm{box} + \mathcal{L}_\mathrm{mask}
\end{align}
Finally, we further finetune UNINEXT on video-level datasets for various downstream object tracking tasks and benchmarks. In this stage, the model is trained on two frames randomly chosen from the original videos. Besides, to avoid the model forgetting previously learned knowledge on image-level tasks, we also transform image-level datasets to pseudo videos for joint training with other video datasets. In summary, the training data in the third stage includes pseudo videos generated from COCO~\cite{COCO}, RefCOCO/g/+~\cite{RefCOCO&plus,RefCOCOg-umd,RefCOCO&plus}, SOT\&VOS datasets (GOT-10K~\cite{GOT10K}, LaSOT~\cite{LaSOT}, TrackingNet~\cite{trackingnet}, and Youtube-VOS~\cite{YoutubeVOS}), MOT\&VIS datasets (BDD100K~\cite{BDD100K}, VIS19~\cite{VIS}, OVIS~\cite{OVIS}), and R-VOS dataset Ref-Youtube-VOS~\cite{URVOS}. Meanwhile, a reference visual encoder for SOT\&VOS and an extra embedding head for association are introduced and optimized in this period. 
\begin{align}
\mathcal{L}_\mathrm{stage3} = \mathcal{L}_\mathrm{retrieve} + \mathcal{L}_\mathrm{box} + \mathcal{L}_\mathrm{mask} + \mathcal{L}_\mathrm{embed}
\end{align}
\textbf{Inference.} For category-guided tasks, UNINEXT predicts instances of different categories and associates them with previous trajectories. The association proceeds in an online fashion and is purely based on the learned instance embedding following~\cite{QDTrack,IDOL}. For expression-guided and annotation-guided tasks, we directly pick the object with the highest matching score with the given prompt as the final result. Different from previous works~\cite{ReferFormer,SiamRCNN} restricted by the offline fashion or complex post-processing, our method is simple, online, and post-processing free.

\section{Experiments}
\subsection{Implementation Details}
We attempt three different backbones, ResNet-50~\cite{ResNet}, ConvNeXt-Large~\cite{ConvNeXt}, and ViT-Huge~\cite{ViT} as the visual encoder. We adopt BERT~\cite{BERT} as the text encoder and its parameters are trained in the first and second training stages while being frozen in the last training stage. The Transformer encoder-decoder architecture follows ~\cite{DeformableDETR} with $6$ encoder layers and $6$ decoder layers. The number of object queries $N$ is set to $900$. The optimizer is AdamW~\cite{AdamW} with weight decay of $0.05$. The model is trained on $32$ and $16$ A100 GPUs for Objects365 pretraining and other stages respectively. More details can be found in the appendix.




\subsection{Evaluations on 10 Tasks}

We compare UNINEXT with task-specific counterparts in 20 datasets. In each benchmark, the best two results are indicated in \rankfirst{bold} and with \ranksecond{underline}. UNINEXT in all benchmarks uses the same model parameters.

\begin{table}[!t]
\caption{State-of-the-art comparison on object detection.}
\label{tab-od}
\vspace{-3mm}
\begin{center}
{\resizebox{\linewidth}{!}{
\begin{tabular}{l|c|cccccc}
\toprule
Model & Backbone & AP & AP$_{50}$ & AP$_{75}$ & AP$_{S}$ & AP$_{M}$ & AP$_{L}$\\
\midrule
Faster R-CNN~\cite{FasterRCNN} & \multirow{7}{*}{ResNet-50} & $42.0$ & $62.1$ & $45.5$ & $26.6$ & $45.4$ & $53.4$ \\
DETR~\cite{DETR} & & $43.3$ & $63.1$ & $45.9$ & $22.5$ & $47.3$ & $61.1$ \\
Sparse R-CNN~\cite{sparsercnn} & & $45.0$ & $63.4$ & $48.2$ & $26.9$ & $47.2$ & $59.5$ \\
Cascade Mask-RCNN~\cite{CascadeRCNN} &  & $46.3$ & $64.3$ & $50.5$ & - & - & - \\
Deformable-DETR~\cite{DeformableDETR}  &  & $46.9$ & $65.6$ & $51.0$ & $29.6$ &  $50.1$ & $61.6$ \\
DN-Deformable-DETR~\cite{DN-DETR}  & & \ranksecond{$48.6$} & \ranksecond{$67.4$} & \ranksecond{$52.7$} & \ranksecond{$31.0$} & \ranksecond{$52.0$} & \ranksecond{$63.7$} \\ 
\textbf{UNINEXT} & & \rankfirst{51.3} & \rankfirst{68.4} & \rankfirst{56.2} & \rankfirst{32.6} & \rankfirst{55.7} & \rankfirst{66.5} \\
\hline
HTC++~\cite{HTC} & \multirow{2}{*}{Swin-L} & 58.0 & - & - & - & - & - \\
DyHead~\cite{DyHead} &  & \ranksecond{60.3} & - & - & - & - & - \\
\hline
Cascade Mask R-CNN~\cite{CascadeRCNN} & \multirow{2}{*}{ConvNeXt-L} & 54.8 & 73.8 & 59.8 & - & - & - \\
\textbf{UNINEXT} &  & 58.1 & \ranksecond{74.9} & \ranksecond{63.7} & \ranksecond{40.7} & \ranksecond{62.5} & \ranksecond{73.6} \\
\hline
ViTDet-H~\cite{ViTDet} & \multirow{2}{*}{ViT-H} & 58.7 & - & - & - & - & - \\
\textbf{UNINEXT} &  & \rankfirst{60.6} & \rankfirst{77.5} & \rankfirst{66.7} & \rankfirst{45.1} & \rankfirst{64.8} & \rankfirst{75.3} \\
\bottomrule
\end{tabular}
}}
\end{center}
\end{table}

\begin{table}[!t]
\vspace{-3mm}
\caption{State-of-the-art comparison on instance segmentation. Methods marked with $^*$ are evaluated on the $\mathtt{val2017}$ split.}
\label{tab-is}
\vspace{-3mm}
\begin{center}
{\resizebox{\linewidth}{!}{
\begin{tabular}{l|c|cccccc}
\toprule
Model & Backbone & AP & AP$_{50}$ & AP$_{75}$ & AP$_{S}$ & AP$_{M}$ & AP$_{L}$\\
\midrule
CondInst~\cite{CondInst} & \multirow{6}{*}{ResNet-50} & $38.6$ & $60.2$ & $41.4$ & $20.6$ & $41.0$ & $51.1$ \\
Cascade Mask R-CNN~\cite{CascadeRCNN} & & $38.6$ & $60.0$ & $41.7$ & $21.7$ & $40.8$ & $49.6$ \\
SOLOv2~\cite{SOLOv2} & & $38.8$ & $59.9$ & $41.7$ & $16.5$ & $41.7$ & \ranksecond{$56.2$} \\
HTC~\cite{HTC} & & $39.7$ & $61.4$ & $43.1$ & $22.6$ & $42.2$ & $50.6$ \\
QueryInst~\cite{QueryInst} &  & \ranksecond{40.6} & \ranksecond{63.0} & \ranksecond{44.0} & \ranksecond{23.4} &  \ranksecond{42.5} & 52.8 \\
\textbf{UNINEXT} & & \rankfirst{44.9} & \rankfirst{67.0} & \rankfirst{48.9} & \rankfirst{26.3} & \rankfirst{48.5} & \rankfirst{59.0} \\
\hline
QueryInst~\cite{QueryInst} & \multirow{2}{*}{Swin-L} & 49.1 & \ranksecond{74.2} & 53.8 & \ranksecond{31.5} & 51.8 & 63.2 \\
Mask2Former~\cite{Mask2Former}$^{*}$ &  & 50.1 & - & - & 29.9 & \ranksecond{53.9} & \rankfirst{72.1} \\
\hline
Cascade Mask R-CNN~\cite{CascadeRCNN} & \multirow{2}{*}{ConvNeXt-L} & 47.6 & 71.3 & 51.7 & - & - & - \\
\textbf{UNINEXT} & & 49.6 & 73.4 & \ranksecond{54.3} & 30.4 & 53.6 & 65.7 \\
\hline
ViTDet-H~\cite{ViTDet}$^{*}$ & \multirow{2}{*}{ViT-H} & \ranksecond{50.9} & - & - & - & - & - \\
\textbf{UNINEXT} &  & \rankfirst{51.8} & \rankfirst{76.2} & \rankfirst{56.7} & \rankfirst{33.3} & \rankfirst{55.9} & \ranksecond{67.5} \\
\bottomrule
\end{tabular}
}}
\end{center}
\vspace{-5mm}
\end{table}

\textbf{Object Detection and Instance Segmentation.} We compare UNINEXT with state-of-the-art object detection and instance segmentation methods on COCO $\mathtt{val2017}$ ($5k$ images) and $\mathtt{test}$-$\mathtt{dev}$ split ($20k$ images) respectively. As shown in Table~\ref{tab-od}, UNINEXT surpasses state-of-the-art query-based detector DN-Deformable DETR~\cite{DN-DETR} by $2.7$ box AP. By replacing ResNet-50~\cite{ResNet} with stronger ConvNeXt-Large~\cite{ConvNeXt} and ViT-Huge~\cite{ViT} backbones, UNINEXT achieves a box AP of $58.1$ and $60.6$, surpassing competitive rivals Cascade Mask-RCNN~\cite{CascadeRCNN} and ViTDet-H~\cite{ViTDet} by $3.3$ and $1.9$ respectively. Besides, the results of instance segmentation are shown in Table~\ref{tab-is}. With the same ResNet-50 backbone, UNINEXT outperforms state-of-the-art QueryInst by $4.3$ AP and $6.2$ AP$_{L}$. When using ConvNeXt-Large as the backbone, UNINEXT achieves a mask AP of $49.6$, surpassing Cascade Mask R-CNN~\cite{CascadeRCNN} by $2.0$. With ViT-Huge as the backbone, UNINEXT achieves state-of-the-art mask AP of $51.8$.

\textbf{REC and RES.} RefCOCO~\cite{RefCOCO&plus}, RefCOCO+~\cite{RefCOCO&plus}, and RefCOCOg~\cite{RefCOCOg-g} are three representative benchmarks for REC and RES proposed by different institutions. Following previous literature, we adopt Precision@0.5 and overall IoU (oIoU) as the evaluation metrics for REC and RES respectively and results are rounded to two decimal places. As shown in Table~\ref{tab-rec} and Table~\ref{tab-res}, our method with ResNet-50 backbone surpasses all previous approaches on all splits. Furthermore, when using ConvNeXt-Large and ViT-Huge backbones, UNINEXT obtains new state-of-the-art results, exceeding the previous best method by a large margin. Especially on RES, UNINEXT-H outperforms LAVT~\cite{LAVT} by $10.85$ on average.

\begin{table}[!t]
\caption{State-of-the-art comparison on REC.}
\label{tab-rec}
\vspace{-3mm}
\begin{center}
{\resizebox{1.0\linewidth}{!}{
\begin{tabular}{lccc|ccc|cc}
\hline\noalign{\smallskip}
\multirow{2}{*}{Method}& \multicolumn{3}{c}{RefCOCO} & \multicolumn{3}{c}{RefCOCO+} & \multicolumn{2}{c}{RefCOCOg}\\
& val & testA & testB & val & testA & testB & val-u & test-u\\
\noalign{\smallskip}
\hline
\noalign{\smallskip}
$\text{UNITER}_{L}$~\cite{Uniter}& 81.41 & 87.04 & 74.17 & 75.90 & 81.45 & 66.70 & 74.86 & 75.77 \\
$\text{VILLA}_{L}$~\cite{VILLA}& 82.39 & 87.48 & 74.84 & 76.17 & 81.54 & 66.84 & 76.18 & 76.71 \\
MDETR~\cite{MDETR}& 86.75 & 89.58 & 81.41 & 79.52 & 84.09 & 70.62 & 81.64 & 80.89 \\
RefTR~\cite{RefTR}& 85.65 & 88.73 & 81.16 & 77.55 & 82.26 & 68.99 & 79.25 & 80.01 \\
SeqTR~\cite{SeqTR}& 87.00 & 90.15 & 83.59 & 78.69 & 84.51 & 71.87 & 82.69 & 83.37 \\
\textbf{UNINEXT-R50}&89.72&91.52& 86.93 & 79.76 & 85.23 & 72.78 & 83.95 & 84.31 \\
\textbf{UNINEXT-L}&\ranksecond{91.43}& \ranksecond{93.73} & \ranksecond{88.93} & \ranksecond{83.09} & \ranksecond{87.90} & \ranksecond{76.15} & \ranksecond{86.91} &\ranksecond{87.48}\\
\textbf{UNINEXT-H}&\rankfirst{92.64}& \rankfirst{94.33} & \rankfirst{91.46} & \rankfirst{85.24} & \rankfirst{89.63} & \rankfirst{79.79} & \rankfirst{88.73} &\rankfirst{89.37}\\
\hline
\end{tabular}
}}
\end{center}
\end{table}

\begin{table}[!t]
\vspace{-3mm}
\caption{State-of-the-art comparison on RES.}
\label{tab-res}
\vspace{-3mm}
\begin{center}
{\resizebox{1.0\linewidth}{!}{
\begin{tabular}{lccc|ccc|cc}
\hline\noalign{\smallskip}
\multirow{2}{*}{Method}& \multicolumn{3}{c}{RefCOCO} & \multicolumn{3}{c}{RefCOCO+} & \multicolumn{2}{c}{RefCOCOg} \\
& val & testA & testB & val & testA & testB & val-u & test-u \\
\noalign{\smallskip}
\hline
\noalign{\smallskip}
CMSA~\cite{CMSA} & 58.32 & 60.61 & 55.09 & 43.76 & 47.60 & 37.89 & - & - \\
BRINet~\cite{BRINet} &60.98 & 62.99 & 59.21 & 48.17 & 52.32 & 42.11 & - & - \\
CMPC+~\cite{CMPC+} & 62.47 & 65.08 & 60.82 & 50.25 & 54.04 & 43.47 & - & - \\
MCN~\cite{MCN} & 62.44 & 64.20 & 59.71 & 50.62 & 54.99 & 44.69 & 49.22 & 49.40 \\
EFN~\cite{EFN} & 62.76 & 65.69 & 59.67 & 51.50 & 55.24 & 43.01 & - & - \\
VLT~\cite{VLT} & 65.65 & 68.29 & 62.73 & 55.50 & 59.20 & 49.36 & 52.99 & 56.65 \\
SeqTR~\cite{SeqTR} &71.70 & 73.31 & 69.82 & 63.04 & 66.73 & 58.97 & 64.69 & 65.74 \\
LAVT~\cite{LAVT} & 72.73 & 75.82 & 68.79 & 62.14 & 68.38 & 55.10 & 61.24 & 62.09 \\
\textbf{UNINEXT-R50} & 77.90 & 79.68 & 75.77 & 66.20 & 71.22 & 59.01 & 70.04 & 70.52 \\
\textbf{UNINEXT-L} & \ranksecond{80.32} & \ranksecond{82.61} & \ranksecond{77.76} & \ranksecond{70.04} & \ranksecond{74.91} & \ranksecond{62.57} & \ranksecond{73.41} & \ranksecond{73.68} \\
\textbf{UNINEXT-H} & \rankfirst{82.19} & \rankfirst{83.44} & \rankfirst{81.33} & \rankfirst{72.47} & \rankfirst{76.42} & \rankfirst{66.22} & \rankfirst{74.67} & \rankfirst{76.37} \\
\noalign{\smallskip}
\hline
\end{tabular}
}}
\end{center}
\vspace{-3mm}
\end{table}


\textbf{SOT.} We compare UNINEXT with state-of-the-art SOT methods on four large-scale benchmarks: LaSOT~\cite{LaSOT}, LaSOT-ext~\cite{lasot_ext}, TrackingNet~\cite{trackingnet}, and TNL-2K~\cite{TNL-2K}. These benchmarks adopt the area under the success curve (AUC), normalized precision (P$_{Norm}$), and precision (P) as the evaluation metrics and include $280$, $150$, $511$, and $700$ videos in the test set respectively. As shown in Table~\ref{tab-sot}, UNINEXT achieves the best results in terms of AUC and P among all trackers with ResNet-50 backbone. Especially on TNL-2K, UNINEXT outperforms the second best method TransT~\cite{TransT} by $5.3$ AUC and $5.8$ P respectively. Besides, UNINEXT with stronger backbones obtains the best AUC on all four benchmarks, exceeding Unicorn~\cite{Unicorn} with the same backbone by $3.9$ on LaSOT.

\begin{table*}[!t]
\caption{State-of-the-art comparison on SOT.}
\label{tab-sot}
\vspace{-3mm}
\begin{center}
{
\begin{center}
\resizebox{0.9\linewidth}{!}{
\begin{tabular}{l|c|ccc|ccc|ccc|cc}
\hline
\multirow{2}{*}{Method} &
  \multirow{2}{*}{Backbone} &
  \multicolumn{3}{c|}{LaSOT~\cite{LaSOT}} &
  \multicolumn{3}{c|}{LaSOT$_{\text{ext}}$~\cite{lasot_ext}} &
  \multicolumn{3}{c|}{TrackingNet~\cite{trackingnet}} &
  \multicolumn{2}{c}{TNL-2K~\cite{TNL-2K}} \\ \cline{3-13} 
                                      &         & AUC  & P$_{Norm}$ & P    & AUC  & P$_{Norm}$ & P    & AUC  & P$_{Norm}$ & P    & AUC   & P \\ \hline
PrDiMP~\cite{PrDiMP}                   & \multirow{5}{*}{ResNet-50} & 59.8 & 68.8 & 60.8 & -    & -          & -    & 75.8 & 81.6 & 70.4 &  47.0  & 45.9       \\
LTMU~\cite{ltmu}                  &  & 57.2 & -          & 57.2    & 41.4    & \ranksecond{49.9}          & \ranksecond{47.3}    & - & -       & - & 48.5 & 47.3 \\
TransT~\cite{TransT}                    &  & 64.9 & 73.8       & 69.0 & -    & -          & -    & \ranksecond{81.4} & \ranksecond{86.7}       & \ranksecond{80.3} & \ranksecond{50.7} & \ranksecond{51.7}        \\
KeepTrack~\cite{keeptrack}             &  & \ranksecond{67.1} & \rankfirst{77.2}       & \ranksecond{70.2} & \ranksecond{48.2} & -          & -    & -    & -          & -    & -    & -           \\
\textbf{UNINEXT}  & &\rankfirst{69.2}&\ranksecond{77.1}&\rankfirst{75.5}&\rankfirst{51.2}&\rankfirst{58.1}&\rankfirst{58.1}&\rankfirst{83.2}&\rankfirst{86.9}&\rankfirst{83.3}&\rankfirst{56.0}&\rankfirst{57.5}\\
\hline
SimTrack~\cite{SimTrack}  &\multirow{3}{*}{ViT-B}&69.3&78.5&-&-&-&-&82.3&-&\rankfirst{86.5}&54.8&53.8\\
OSTrack~\cite{OSTrack} &  &
  71.1 &
  \rankfirst{81.1} &
  77.6 &
  50.5 &
  61.3 &
  57.6 &
  83.9 &
  88.5 &
  83.2 &
  55.9 &
  - \\ 
SeqTrack~\cite{SeqTrack} & &
71.5 & 
\rankfirst{81.1} &
77.8 &
50.5 &
61.6 &
57.5 &
83.9 &
\ranksecond{88.8} &
83.6 &
57.8 &
- \\
\hline
Unicorn~\cite{Unicorn}  &\multirow{2}{*}{ConvNeXt-L}&68.5&76.6&74.1&-&-&-&83.0&86.4&82.2&-&-\\
\textbf{UNINEXT}  & &\rankfirst{72.4}&\ranksecond{80.7}&\ranksecond{78.9}&\ranksecond{54.4}&\ranksecond{61.8}&\ranksecond{61.4}&\ranksecond{85.1}&88.2&84.7&\ranksecond{58.1}&\ranksecond{60.7}\\
\hline
\textbf{UNINEXT}  & ViT-H&\ranksecond{72.2}&\ranksecond{80.7}&\rankfirst{79.4}&\rankfirst{56.2}&\rankfirst{63.8}&\rankfirst{63.8}&\rankfirst{85.4}&\rankfirst{89.0}&\ranksecond{86.4}&\rankfirst{59.3}&\rankfirst{62.8}\\
\hline
 \end{tabular}}
\end{center}


}
\end{center}
\end{table*}

\textbf{VOS.} The comparisons between UNINEXT with previous semi-supervised VOS methods are demonstrated in Table~\ref{tab:vos}. DAVIS-2017~\cite{DAVIS17} adopts region similarity \mj, contour accuracy \mf, and the averaged score \mjf\ as the metrics. Similarly, Youtube-VOS $2018$~\cite{YoutubeVOS} reports \mj\ and \mf\ for both seen and unseen categories, and the averaged overall score \mg. UNINEXT achieves the best results among all non-memory-based methods, largely bridging the performance gap between non-memory-based approaches and memory-based ones. Furthermore, compared with traditional memory-based methods~\cite{STM, STCN}, UNINEXT does not rely on the intermediate mask predictions. This leads to constant memory consumption, enabling UNINEXT to handle long sequences of any length.  

\begin{table}[!t]
\vspace{-5mm}
\caption{State-of-the-art comparison on VOS.}
\centering{}{\footnotesize{}}%
{\resizebox{1.0\linewidth}{!}{
\setlength{\tabcolsep}{2pt}
\begin{tabular}{llcccccccc}
\toprule 
\multirow{2}{*}{\ \ \ \ }&\multirow{2}{*}{Method} & \multicolumn{5}{c}{YT-VOS 2018 val~\cite{YoutubeVOS}}  & \multicolumn{3}{c}{DAVIS 2017 val~\cite{DAVIS17}} \\
\cmidrule(lr){3-7} \cmidrule(lr){8-10}
& & \mg & \mjs & \mfs & \mju & \mfu & \mjf & \mj & \mf\\

\midrule

\parbox[t]{2mm}{\multirow{4}{*}{\rotatebox[origin=c]{90}{Memory}}}
&STM~\cite{STM}&79.4&79.7&84.2&72.8&80.9&81.8&79.2&84.3\tabularnewline
&CFBI~\cite{CFBI}&81.4&81.1&85.8&75.3&83.4&81.9&79.1&84.6\tabularnewline
&STCN~\cite{STCN}&83.0&81.9&86.5&77.9&85.7&85.4&82.2&88.6\tabularnewline
&XMem~\cite{XMem}&86.1&85.1&89.8&80.3&89.2&87.7&84.0&91.4\tabularnewline
\midrule

\parbox{2mm}{\multirow{7}{*}{\rotatebox[origin=c]{90}{Non-Memory}}}
&SiamMask~\cite{SiamMask}&52.8&60.2&58.2&45.1&47.7&56.4&54.3&58.5\tabularnewline
&Unicorn~\cite{Unicorn}&-&-&-&-&-&69.2&65.2&73.2\tabularnewline
&Siam R-CNN~\cite{SiamRCNN}&73.2&73.5&-&66.2&-&70.6&66.1&75.0\tabularnewline
&TVOS~\cite{TVOS}&67.8&67.1&69.4&63.0&71.6&72.3&69.9&74.7\tabularnewline
&FRTM~\cite{FRTM}&72.1&72.3&76.2&65.9&74.1&76.7&\ranksecond{73.9}&79.6\tabularnewline
&\textbf{UNINEXT-R50}& 77.0 & 76.8 & 81.0 & \ranksecond{70.8} & \rankfirst{79.4} & 74.5 & 71.3 & 77.6 \tabularnewline
&\textbf{UNINEXT-L}& \ranksecond{78.1} & \ranksecond{79.1} & \ranksecond{83.5} & \rankfirst{71.0} & 78.9 & \ranksecond{77.2} & 73.2 & \ranksecond{81.2} \tabularnewline
&\textbf{UNINEXT-H}& \rankfirst{78.6} & \rankfirst{79.9} & \rankfirst{84.9} & 70.6 & \ranksecond{79.2} & \rankfirst{81.8} & \rankfirst{77.7} & \rankfirst{85.8} \tabularnewline
\bottomrule
\end{tabular}}}
\label{tab:vos}
\vspace{-3mm}
\end{table}

\textbf{MOT.} We compare UNINEXT with state-of-the-art MOT methods on BDD100K~\cite{BDD100K}, which requires tracking $8$ classes of instances in the autonomous driving scenario. Except for classical evaluation metrics Multiple-Object Tracking Accuracy (MOTA), Identity F1 Score (IDF1), and Identity Switches (IDS), BDD100K additionally introduces mMOTA, and mIDF1 to evaluate the average performance across $8$ classes. As shown in Table~\ref{tab:bdd}, UNINEXT surpasses Unicorn~\cite{Unicorn} by $3.0$ mMOTA and $2.7$ mIDF1 respectively.

\textbf{MOTS.} Similar to MOT, BDD100K MOTS Challenge~\cite{BDD100K} evaluates the performance on multi-class tracking by mMOTSA, mMOTSP, mIDF1, and ID Sw. This benchmark contains $37$ sequences with mask annotations in the validation set. As shown in Table~\ref{tab:mots}, UNINEXT achieves state-of-the-art performance, surpassing the previous best method Unicorn~\cite{Unicorn} by $6.1$ mMOTSA.

\textbf{VIS.} We compare UNINEXT against state-of-the-art VIS methods on Youtube-VIS $2019$~\cite{VIS} and OVIS~\cite{OVIS} validation sets. Specifically, Youtube-VIS $2019$ and OVIS have $40$ and $25$ object categories, containing $302$ and $140$ videos respectively in the validation set. Both benchmarks take AP as the main metric. As shown in Table~\ref{tab:vis}, when using the same ResNet-50 backbone,  
UNINEXT obtains the best results on both datasets. Especially on more challenging OVIS, UNINEXT exceeds the previous best method IDOL~\cite{IDOL} by $3.8$ AP. When using stronger ViT-Huge backbone, UNINEXT achieves state-of-the-art AP of $66.9$ on Youtube-VIS $2019$ and $49.0$ on OVIS respectively, surpassing previous methods by a large margin. 


\begin{table}[t]
\vspace{-5mm}
    \caption{State-of-the-art comparison on MOT.}
    \centering
    {    \resizebox{\linewidth}{!}{
        \begin{tabular}{lcccccc}
            \toprule
            Method                  & Split & mMOTA$\uparrow$ & mIDF1$\uparrow$ & MOTA$\uparrow$ & IDF1$\uparrow$ & ID Sw.$\downarrow$ \\
            \midrule
            Yu~\etal~\cite{BDD100K} & val   & 25.9             & 44.5             & 56.9            & 66.8          & \ranksecond{8315}           \\
            QDTrack~\cite{QDTrack}                    & val   & 36.6    & 50.8    & 63.5   & \rankfirst{71.5}   & \rankfirst{6262} \\
            Unicorn~\cite{Unicorn} & val &41.2&54.0&\ranksecond{66.6}&\ranksecond{71.3}&10876\\
            \textbf{UNINEXT-L} & val &\ranksecond{41.8}&\ranksecond{54.9}&64.6&68.7&9134\\
            \textbf{UNINEXT-H} & val &\rankfirst{44.2}&\rankfirst{56.7}&\rankfirst{67.1}&69.9&10222\\
            \bottomrule
        \end{tabular}
    }

}
    \label{tab:bdd}
\end{table}

\begin{table}[t]
    \caption{State-of-the-art comparison on MOTS.}
    \centering
    {    \resizebox{1.0\linewidth}{!}{
        \begin{tabular}{lcccccc}
            \toprule
            Method & Online & mMOTSA$\uparrow$ & mMOTSP$\uparrow$ & mIDF1$\uparrow$ & ID Sw.$\downarrow$ \\
            \midrule
            SortIoU&\ding{51}&10.3&59.9&21.8&15951\\
            MaskTrackRCNN~\cite{VIS}&\ding{51}&12.3&59.9&26.2&9116\\
            STEm-Seg~\cite{stemseg}&\ding{55}&12.2&58.2&25.4&8732\\
            QDTrack-mots~\cite{QDTrack}&\ding{51}&22.5&59.6&40.8&1340\\
            PCAN~\cite{PCAN}&\ding{51}&27.4&66.7&45.1&\ranksecond{876}\\
            VMT~\cite{VMT}&\ding{55}&28.7&67.3&\ranksecond{45.7}&\rankfirst{825}\\
            Unicorn~\cite{Unicorn}&\ding{51}&29.6&\ranksecond{67.7}&44.2&1731\\
            \textbf{UNINEXT-L}&\ding{51}&\ranksecond{32.0}&60.2&45.4&1634\\
            \textbf{UNINEXT-H}&\ding{51}&\rankfirst{35.7}&\rankfirst{68.1}&\rankfirst{48.5}&1776\\
            \bottomrule
        \end{tabular}
    }}
    \label{tab:mots}
\vspace{-3mm}
\end{table}

\textbf{R-VOS.} Ref-Youtube-VOS~\cite{URVOS} and Ref-DAVIS17~\cite{ref-davis} are two popular R-VOS benchmarks, which are constructed by introducing language expressions for the objects in the original Youtube-VOS~\cite{YoutubeVOS} and DAVIS17~\cite{DAVIS17} datasets. As same as semi-supervised VOS, region similarity \mj , contour accuracy \mf, and the averaged score \mjf\ are adopted as the metrics. As demonstrated in Table~\ref{tab:rvos}, UNINEXT outperforms all previous R-VOS approaches by a large margin, when using the same ResNet-50 backbone. Especially on Ref-DAVIS17, UNINEXT exceeds previous best ReferFormer~\cite{ReferFormer} by $5.4$ \mjf. Furthermore, when adopting stronger ViT-Huge backbone, UNINEXT achieves new state-of-the-art \mjf of $70.1$ on Ref-Youtube-VOS and $72.5$ on Ref-DAVIS17. Besides, different from offline RefFormer, UNINEXT works in a flexible online fashion, making it applicable to ongoing videos in the real world.

\begin{table}[t]
    \caption{State-of-the-art comparison on VIS.}
    \centering
    {\resizebox{1.0\columnwidth}{!}{
\begin{tabular}{l|c|c|ccc|ccc}
\hline
\multirow{2}{*}{Method}   & \multirow{2}{*}{Backbone} &\multirow{2}{*}{Online} &\multicolumn{3}{c}{VIS2019 val}&\multicolumn{3}{c}{OVIS val} \\
\arrayrulecolor{white}\cline{4-9}
\arrayrulecolor{black}\cline{4-9}
\arrayrulecolor{black}\cline{4-9}
\arrayrulecolor{black}\cline{4-9}
\arrayrulecolor{white}\cline{4-9}
&&&$\rm AP$    &$\rm AP_{50}$  &$\rm AP_{75}$ &$\rm AP$    &$\rm AP_{50}$  &$\rm AP_{75}$\\
\arrayrulecolor{white}\hline
\arrayrulecolor{black}\hline
\arrayrulecolor{white}\hline

 VisTR~\cite{VISTR}   & \multirow{7}{*}{ResNet-50}      &\ding{55} &36.2 &59.8  &36.9 & - & - & -\\
 MaskProp~\cite{MaskProp}  &    &\ding{55} &40.0 &-&42.9 & - & - & -\\  
 IFC~\cite{IFC}  &   &\ding{55} &42.8 &65.8 &46.8&13.1&27.8&11.6 \\  
{SeqFormer}~\cite{SeqFormer}   &    &\ding{55} &47.4 &69.8 &51.8&15.1&31.9&13.8\\
 IDOL~\cite{IDOL}  &  &\ding{51} & 49.5 &\ranksecond{74.0} &52.9&\ranksecond{30.2}&\ranksecond{51.3}&\ranksecond{30.0}\\
 VITA~\cite{VITA}&&\ding{55}&\ranksecond{49.8}&72.6&\ranksecond{54.5}&19.6&41.2&17.4\\
\textbf{UNINEXT}  &  &\ding{51} &\rankfirst{53.0}& \rankfirst{75.2}& \rankfirst{59.1} & \rankfirst{34.0} & \rankfirst{55.5} & \rankfirst{35.6}\\  

\arrayrulecolor{white}\hline
\arrayrulecolor{black}\hline
\arrayrulecolor{white}\hline
 
  SeqFormer~\cite{SeqFormer}  &  \multirow{4}{*}{Swin-L} &\ding{55} &{59.3}  &{82.1}  &{66.4} & - & - & -\\ 
  VMT~\cite{VMT}&&\ding{55}&59.7&-&66.7&19.8&39.6&17.2\\
 VITA~\cite{VITA}&&\ding{55}&63.0&86.9&67.9&-&-&-\\
   IDOL~\cite{IDOL}  &  &\ding{51} &\ranksecond{64.3} &\rankfirst{87.5} &71.0&\ranksecond{42.6}&65.7&\ranksecond{45.2}\\  
\arrayrulecolor{white}\hline
\arrayrulecolor{black}\hline
\arrayrulecolor{white}\hline
   \textbf{UNINEXT}  & ConvNeXt-L &\ding{51} &\ranksecond{64.3}&\ranksecond{87.2}&\ranksecond{71.7}&41.1&\ranksecond{65.8}&42.0\\  
   \textbf{UNINEXT}  & ViT-H &\ding{51} &\rankfirst{66.9}&\rankfirst{87.5}&\rankfirst{75.1}&\rankfirst{49.0}&\rankfirst{72.5}&\rankfirst{52.2}\\  
\arrayrulecolor{white}\hline
\arrayrulecolor{black}\hline
\arrayrulecolor{white}\hline
\end{tabular}
}}
    \label{tab:vis}
\end{table}

\begin{table}[t]
    \caption{State-of-the-art comparison on R-VOS.}
    \centering
    {\resizebox{1.0\linewidth}{!}{
\begin{tabular}{l | c | c c c | c c c}

\toprule

\multirow{2}{*}{Method} & \multirow{2}{*}{Backbone} & \multicolumn{3}{c}{Ref-Youtube-VOS} & \multicolumn{3}{c}{Ref-DAVIS17} \\

\arrayrulecolor{white}\cline{3-8}
\arrayrulecolor{black}\cline{3-8}
\arrayrulecolor{black}\cline{3-8}
\arrayrulecolor{black}\cline{3-8}
\arrayrulecolor{white}\cline{3-8}

 & & $\mathcal{J}\&\mathcal{F}$ & $\mathcal{J}$ & $\mathcal{F}$ & 
     $\mathcal{J}\&\mathcal{F}$ & $\mathcal{J}$ & $\mathcal{F}$ \\

\arrayrulecolor{white}\hline
\arrayrulecolor{black}\hline
\arrayrulecolor{white}\hline

CMSA~\cite{CMSA}  &\multirow{5}{*}{ResNet-50}& 36.4 & 34.8 & 38.1 & 40.2 & 36.9 & 43.5 \\ 
URVOS~\cite{URVOS} &  & 47.2 & 45.3 & 49.2 & 51.5 & 47.3 & 56.0 \\
YOFO~\cite{YOFO}& &48.6&47.5&49.7&54.4&50.1&58.7\\
ReferFormer~\cite{ReferFormer} &  & \ranksecond{58.7} & \ranksecond{57.4} & \ranksecond{60.1} & \ranksecond{58.5} & \ranksecond{55.8} & \ranksecond{61.3} \\
\textbf{UNINEXT}& & \rankfirst{61.2} & \rankfirst{59.3}& \rankfirst{63.0} & \rankfirst{63.9} & \rankfirst{59.6} & \rankfirst{68.1} \\

\arrayrulecolor{white}\hline
\arrayrulecolor{black}\hline
\arrayrulecolor{white}\hline

PMINet + CFBI ~\cite{PMINet} & \multirow{2}{*}{Ensemble} & 54.2 & 53.0 & 55.5 & - & - & - \\
CITD ~\cite{CITD} &  & 61.4 & 60.0 & 62.7 & - & - & - \\
\arrayrulecolor{white}\hline
\arrayrulecolor{black}\hline
\arrayrulecolor{white}\hline
MTTR~\cite{MTTR} & \multirow{2}{*}{Video-Swin-T} & 55.3 & 54.0 & 56.6 & - & - & - \\
ReferFormer~\cite{ReferFormer} &  & 64.9 & 62.8 & 67.0 & 61.1 & 58.1 & 64.1 \\
\arrayrulecolor{white}\hline
\arrayrulecolor{black}\hline
\arrayrulecolor{white}\hline
\textbf{UNINEXT}&ConvNext-L&\ranksecond{66.2}&\ranksecond{64.0}&\ranksecond{68.4}&\ranksecond{66.7}&\ranksecond{62.3}&\ranksecond{71.1}\\
\textbf{UNINEXT}&ViT-H&\rankfirst{70.1} & \rankfirst{67.6} & \rankfirst{72.7}&\rankfirst{72.5}&\rankfirst{68.2}&\rankfirst{76.8}\\
\arrayrulecolor{white}\hline
\arrayrulecolor{black}\hline
\arrayrulecolor{white}\hline

\end{tabular}
}}
    \label{tab:rvos}
\vspace{-3mm}
\end{table}

\subsection{Ablations and Other Analysis}
\label{sec:ablation}

In this section, we conduct component-wise analysis for better understanding our method. All models take ResNet-50 as the backbone. The methods are evaluated on five benchmarks (COCO~\cite{COCO}, RefCOCO~\cite{RefCOCO&plus}, Youtube-VOS~\cite{YoutubeVOS}, Ref-Youtube-VOS~\cite{URVOS}, and Youtube-VIS $2019$~\cite{VIS}) from five tasks (object detection, REC, VOS, R-VOS, and VIS). The results are shown in Table~\ref{tab:ablation}. 

\textbf{Fusion.} To study the effect of feature fusion between visual features and prompt embeddings, we implement a variant without any early fusion. In this version, prompt embeddings do not have an influence on proposal generation but are only used in the final object retrieval process. Experiments show that early fusion has the greatest impact on VOS, the performance on VOS drops drastically by $21.4$ \mjf\ without feature fusion. This is mainly caused by the following reasons (1) Without the guidance of prompt embeddings, the network can hardly find rare referred targets like trees and sinks. (2) Without early fusion, the network cannot fully exploit fine mask annotations in the first frame, causing degradation of the mask quality. Besides, the removal of feature fusion also causes performance drop of $2.3$ P@0.5 and $2.8$ \mjf on REC and RVOS respectively, showing the importance of early fusion in expression-guided tasks. Finally, feature fusion has minimum influence on object detection and VIS. This can be understood because both two tasks aim to find all objects as completely as possible rather than locating one specific target referred by the prompt. 

\textbf{Queries.} We compare two different query generation strategies: static queries by \texttt{nn.Embedding(N, d)} and dynamic queries conditioned on the prompt embeddings. Experiments show that dynamic queries perform slightly better than static queries on the first four tasks. However, static queries outperform dynamic ones by 2.8 AP on the VIS task, obtaining higher overall performance. A potential reason is that $N$ different object queries can encode richer inner relationship among different targets than simply copying the pooled prompt by $N$ times as queries. This is especially important for VIS because targets need to be associated according to their affinity in appearance and space.

\textbf{Unification.} We also compare two different model design philosophies, one unified model or multiple task-specific models. Except for the unified model, we also retrain five task-specific models only on data from corresponding tasks. Experiments show that the unified model achieves significantly better performance than its task-specific counterparts on five tasks, demonstrating the superiority of the unified formulation and joint training on all instance perception tasks. 
Finally, the unified model can save tons of parameters, being much more parameter-efficient. 

\begin{table}[t]
    \caption{Ablations. The settings in our final model is underlined.}
    \centering
    {\resizebox{1.0\linewidth}{!}{
\begin{tabular}{clccccc}
\toprule
\multirow{3}{*}{Experiment} & \multirow{3}{*}{Method} & \multicolumn{1}{c}{\underline{OD}} & 
\multicolumn{1}{c}{\underline{REC}} & 
\multicolumn{1}{c}{\underline{VOS}} & 
\multicolumn{1}{c}{\underline{RVOS}} & 
\multicolumn{1}{c}{\underline{VIS}} \\
& &COCO&RefCOCO&YTBVOS&R-YTBVOS&VIS19  \\
& &(AP)&(P@0.5)&($\mathcal{J\&F}$)&($\mathcal{J\&F}$)&(AP)\\ 
\midrule 
\midrule
\multirow{3}{*}{Fusion}  & \underline{Early Fusion} &51.3&89.7&77.0&61.2&53.0  \\
                            & W/o Fusion&51.1&87.4& 55.6 & 58.4 &51.0 \\ 
                            & & (+0.2) & (+2.3) & (+21.4) & (+2.8) & (+2.0) \\ \midrule
\multirow{3}{*}{Queries}      
    &\underline{Static}&51.3&89.7&77.0&61.2&53.0  \\
    & Dynamic&51.9&89.8&77.4&61.6&50.2 \\ 
    & & (-0.6) & (-0.1) & (-0.4) & (-0.4) & (+2.8) \\
                            \midrule
\multirow{3}{*}{Model}    
&\underline{Unified}&51.3&89.7&77.0&61.2&53.0  \\
&Task-specific& 50.8 & 87.6 & 74.2 & 57.2 & 50.1 \\
& & (+0.5) & (+2.1) & (+2.8) & (+4.0) & (+2.9) \\
\midrule
\end{tabular}
}}
    \label{tab:ablation}
    \vspace{-3mm}
\end{table}

\section{Conclusions}
We propose UNINEXT, a universal instance perception model of the next generation. For the first time, UNINEXT unifies 10 instance perception tasks with a prompt-guided object discovery and retrieval paradigm. Extensive experiments demonstrate that UNINEXT achieves superior performance on 20 challenging benchmarks with a single model with the same model parameters. We hope that UNINEXT can serve as a solid baseline for the research of instance perception in the future.

\noindent{\textbf{Acknowledgement.}
We would like to thank the reviewers for their insightful comments. The paper is supported in part by the National Key R\&D Program of China under Grant No. 2018AAA0102001, 2022ZD0161000 and National Natural Science Foundation of China under grant No. 62293542, U1903215, 62022021 and the Fundamental Research Funds for the Central Universities No.DUT22ZD210.}
{\small
\bibliographystyle{ieee_fullname}
\bibliography{egbib}

\begin{thebibliography}{100}\itemsep=-1pt

\bibitem{stemseg}
Ali Athar, Sabarinath Mahadevan, Aljosa Osep, Laura Leal-Taix{\'e}, and Bastian
  Leibe.
\newblock {STEm-Seg}: Spatio-temporal embeddings for instance segmentation in
  videos.
\newblock In {\em ECCV}, 2020.

\bibitem{Pathways}
Paul Barham, Aakanksha Chowdhery, Jeff Dean, Sanjay Ghemawat, Steven Hand,
  Daniel Hurt, Michael Isard, Hyeontaek Lim, Ruoming Pang, Sudip Roy, et~al.
\newblock Pathways: Asynchronous distributed dataflow for {ML}.
\newblock {\em PMLS}, 2022.

\bibitem{Tracktor}
Philipp Bergmann, Tim Meinhardt, and Laura Leal-Taixe.
\newblock Tracking without bells and whistles.
\newblock In {\em ICCV}, 2019.

\bibitem{MaskProp}
Gedas Bertasius and Lorenzo Torresani.
\newblock Classifying, segmenting, and tracking object instances in video with
  mask propagation.
\newblock In {\em CVPR}, 2020.

\bibitem{SiameseFC}
Luca Bertinetto, Jack Valmadre, Jo{\~a}o~F Henriques, Andrea Vedaldi, and
  Philip H~S Torr.
\newblock Fully-convolutional siamese networks for object tracking.
\newblock In {\em ECCVW}, 2016.

\bibitem{YOLACT}
Daniel Bolya, Chong Zhou, Fanyi Xiao, and Yong~Jae Lee.
\newblock {YOLACT}: Real-time instance segmentation.
\newblock In {\em ICCV}, 2019.

\bibitem{MTTR}
Adam Botach, Evgenii Zheltonozhskii, and Chaim Baskin.
\newblock End-to-end referring video object segmentation with multimodal
  transformers.
\newblock In {\em CVPR}, 2022.

\bibitem{CascadeRCNN}
Zhaowei Cai and Nuno Vasconcelos.
\newblock {Cascade R-CNN}: high quality object detection and instance
  segmentation.
\newblock {\em TPAMI}, 2019.

\bibitem{DETR}
Nicolas Carion, Francisco Massa, Gabriel Synnaeve, Nicolas Usunier, Alexander
  Kirillov, and Sergey Zagoruyko.
\newblock End-to-end object detection with transformers.
\newblock In {\em ECCV}, 2020.

\bibitem{SimTrack}
Boyu Chen, Peixia Li, Lei Bai, Lei Qiao, Qiuhong Shen, Bo Li, Weihao Gan, Wei
  Wu, and Wanli Ouyang.
\newblock Backbone is all your need: A simplified architecture for visual
  object tracking.
\newblock In {\em ECCV}, 2022.

\bibitem{STEP}
Ding-Jie Chen, Songhao Jia, Yi-Chen Lo, Hwann-Tzong Chen, and Tyng-Luh Liu.
\newblock See-through-text grouping for referring image segmentation.
\newblock In {\em ICCV}, 2019.

\bibitem{HTC}
Kai Chen, Jiangmiao Pang, Jiaqi Wang, Yu Xiong, Xiaoxiao Li, Shuyang Sun,
  Wansen Feng, Ziwei Liu, Jianping Shi, Wanli Ouyang, et~al.
\newblock Hybrid task cascade for instance segmentation.
\newblock In {\em CVPR}, 2019.

\bibitem{Pix2SeqV2}
Ting Chen, Saurabh Saxena, Lala Li, Tsung-Yi Lin, David~J Fleet, and Geoffrey
  Hinton.
\newblock A unified sequence interface for vision tasks.
\newblock {\em NeurIPS}, 2022.

\bibitem{SeqTrack}
Xin Chen, Houwen Peng, Dong Wang, Huchuan Lu, and Han Hu.
\newblock {SeqTrack}: Sequence to sequence learning for visual object tracking.
\newblock In {\em CVPR}, 2023.

\bibitem{TransT}
Xin Chen, Bin Yan, Jiawen Zhu, Dong Wang, Xiaoyun Yang, and Huchuan Lu.
\newblock Transformer tracking.
\newblock In {\em CVPR}, 2021.

\bibitem{Uniter}
Yen-Chun Chen, Linjie Li, Licheng Yu, Ahmed El~Kholy, Faisal Ahmed, Zhe Gan, Yu
  Cheng, and Jingjing Liu.
\newblock Uniter: Universal image-text representation learning.
\newblock In {\em ECCV}, 2020.

\bibitem{Mask2Former}
Bowen Cheng, Ishan Misra, Alexander~G Schwing, Alexander Kirillov, and Rohit
  Girdhar.
\newblock Masked-attention mask transformer for universal image segmentation.
\newblock In {\em CVPR}, 2022.

\bibitem{XMem}
Ho~Kei Cheng and Alexander~G Schwing.
\newblock Xmem: Long-term video object segmentation with an atkinson-shiffrin
  memory model.
\newblock In {\em ECCV}, 2022.

\bibitem{STCN}
Ho~Kei Cheng, Yu-Wing Tai, and Chi-Keung Tang.
\newblock Rethinking space-time networks with improved memory coverage for
  efficient video object segmentation.
\newblock {\em NeurIPS}, 2021.

\bibitem{MixFormer}
Yutao Cui, Cheng Jiang, Limin Wang, and Gangshan Wu.
\newblock Mixformer: End-to-end tracking with iterative mixed attention.
\newblock In {\em CVPR}, 2022.

\bibitem{ltmu}
Kenan Dai, Yunhua Zhang, Dong Wang, Jianhua Li, Huchuan Lu, and Xiaoyun Yang.
\newblock High-performance long-term tracking with meta-updater.
\newblock In {\em CVPR}, 2020.

\bibitem{DyHead}
Xiyang Dai, Yinpeng Chen, Bin Xiao, Dongdong Chen, Mengchen Liu, Lu Yuan, and
  Lei Zhang.
\newblock Dynamic head: Unifying object detection heads with attentions.
\newblock In {\em CVPR}, 2021.

\bibitem{PrDiMP}
Martin Danelljan, Luc~Van Gool, and Radu Timofte.
\newblock Probabilistic regression for visual tracking.
\newblock In {\em CVPR}, 2020.

\bibitem{TAO}
Achal Dave, Tarasha Khurana, Pavel Tokmakov, Cordelia Schmid, and Deva Ramanan.
\newblock {TAO}: A large-scale benchmark for tracking any object.
\newblock In {\em ECCV}, 2020.

\bibitem{TransVG}
Jiajun Deng, Zhengyuan Yang, Tianlang Chen, Wengang Zhou, and Houqiang Li.
\newblock {TransVG}: End-to-end visual grounding with transformers.
\newblock In {\em ICCV}, 2021.

\bibitem{BERT}
Jacob Devlin, Ming-Wei Chang, Kenton Lee, and Kristina Toutanova.
\newblock {BERT}: Pre-training of deep bidirectional transformers for language
  understanding.
\newblock {\em NAACL-HLT}, 2019.

\bibitem{VLT}
Henghui Ding, Chang Liu, Suchen Wang, and Xudong Jiang.
\newblock Vision-language transformer and query generation for referring
  segmentation.
\newblock In {\em ICCV}, 2021.

\bibitem{PMINet}
Zihan Ding, Tianrui Hui, Shaofei Huang, Si Liu, Xuan Luo, Junshi Huang, and
  Xiaoming Wei.
\newblock Progressive multimodal interaction network for referring video object
  segmentation.
\newblock {\em The 3rd Large-scale Video Object Segmentation Challenge}, 2021.

\bibitem{ViT}
Alexey Dosovitskiy, Lucas Beyer, Alexander Kolesnikov, Dirk Weissenborn,
  Xiaohua Zhai, Thomas Unterthiner, Mostafa Dehghani, Matthias Minderer, Georg
  Heigold, Sylvain Gelly, et~al.
\newblock An image is worth 16x16 words: Transformers for image recognition at
  scale.
\newblock {\em arXiv preprint arXiv:2010.11929}, 2020.

\bibitem{lasot_ext}
Heng Fan, Hexin Bai, Liting Lin, Fan Yang, Peng Chu, Ge Deng, Sijia Yu,
  Mingzhen Huang, Juehuan Liu, Yong Xu, et~al.
\newblock {LaSOT}: A high-quality large-scale single object tracking benchmark.
\newblock {\em IJCV}, 2021.

\bibitem{LaSOT}
Heng Fan, Liting Lin, Fan Yang, Peng Chu, Ge Deng, Sijia Yu, Hexin Bai, Yong
  Xu, Chunyuan Liao, and Haibin Ling.
\newblock {LaSOT}: A high-quality benchmark for large-scale single object
  tracking.
\newblock In {\em CVPR}, 2019.

\bibitem{QueryInst}
Yuxin Fang, Shusheng Yang, Xinggang Wang, Yu Li, Chen Fang, Ying Shan, Bin
  Feng, and Wenyu Liu.
\newblock Instances as queries.
\newblock In {\em ICCV}, 2021.

\bibitem{DPM}
Pedro~F Felzenszwalb, Ross~B Girshick, David McAllester, and Deva Ramanan.
\newblock Object detection with discriminatively trained part-based models.
\newblock {\em TPAMI}, 2010.

\bibitem{EFN}
Guang Feng, Zhiwei Hu, Lihe Zhang, and Huchuan Lu.
\newblock Encoder fusion network with co-attention embedding for referring
  image segmentation.
\newblock In {\em CVPR}, 2021.

\bibitem{VILLA}
Zhe Gan, Yen-Chun Chen, Linjie Li, Chen Zhu, Yu Cheng, and Jingjing Liu.
\newblock Large-scale adversarial training for vision-and-language
  representation learning.
\newblock {\em NeurIPS}, 2020.

\bibitem{YOLOX}
Zheng Ge, Songtao Liu, Feng Wang, Zeming Li, and Jian Sun.
\newblock {YOLOX}: Exceeding yolo series in 2021.
\newblock {\em arXiv preprint arXiv:2107.08430}, 2021.

\bibitem{MuST}
Golnaz Ghiasi, Barret Zoph, Ekin~D Cubuk, Quoc~V Le, and Tsung-Yi Lin.
\newblock Multi-task self-training for learning general representations.
\newblock In {\em ICCV}, 2021.

\bibitem{FastRCNN}
Ross Girshick.
\newblock {Fast R-CNN}.
\newblock In {\em ICCV}, 2015.

\bibitem{LVIS}
Agrim Gupta, Piotr Dollar, and Ross Girshick.
\newblock {LVIS}: A dataset for large vocabulary instance segmentation.
\newblock In {\em CVPR}, 2019.

\bibitem{MaskRCNN}
Kaiming He, Georgia Gkioxari, Piotr Doll{\'a}r, and Ross Girshick.
\newblock Mask {R-CNN}.
\newblock In {\em ICCV}, 2017.

\bibitem{ResNet}
Kaiming He, Xiangyu Zhang, Shaoqing Ren, and Jian Sun.
\newblock Deep residual learning for image recognition.
\newblock In {\em CVPR}, 2016.

\bibitem{VITA}
Miran Heo, Sukjun Hwang, Seoung~Wug Oh, Joon-Young Lee, and Seon~Joo Kim.
\newblock Vita: Video instance segmentation via object token association.
\newblock {\em NeurIPS}, 2022.

\bibitem{RvG-Tree}
Richang Hong, Daqing Liu, Xiaoyu Mo, Xiangnan He, and Hanwang Zhang.
\newblock Learning to compose and reason with language tree structures for
  visual grounding.
\newblock {\em TPAMI}, 2019.

\bibitem{BRINet}
Zhiwei Hu, Guang Feng, Jiayu Sun, Lihe Zhang, and Huchuan Lu.
\newblock Bi-directional relationship inferring network for referring image
  segmentation.
\newblock In {\em CVPR}, 2020.

\bibitem{GOT10K}
Lianghua Huang, Xin Zhao, and Kaiqi Huang.
\newblock {GOT-10k}: A large high-diversity benchmark for generic object
  tracking in the wild.
\newblock {\em TPAMI}, 2019.

\bibitem{IFC}
Sukjun Hwang, Miran Heo, Seoung~Wug Oh, and Seon~Joo Kim.
\newblock Video instance segmentation using inter-frame communication
  transformers.
\newblock {\em NeurIPS}, 2021.

\bibitem{LTS}
Ya Jing, Tao Kong, Wei Wang, Liang Wang, Lei Li, and Tieniu Tan.
\newblock Locate then segment: A strong pipeline for referring image
  segmentation.
\newblock In {\em CVPR}, 2021.

\bibitem{MDETR}
Aishwarya Kamath, Mannat Singh, Yann LeCun, Gabriel Synnaeve, Ishan Misra, and
  Nicolas Carion.
\newblock Mdetr-modulated detection for end-to-end multi-modal understanding.
\newblock In {\em ICCV}, 2021.

\bibitem{VMT}
Lei Ke, Henghui Ding, Martin Danelljan, Yu-Wing Tai, Chi-Keung Tang, and Fisher
  Yu.
\newblock Video mask transfiner for high-quality video instance segmentation.
\newblock {\em ECCV}, 2022.

\bibitem{PCAN}
Lei Ke, Xia Li, Martin Danelljan, Yu-Wing Tai, Chi-Keung Tang, and Fisher Yu.
\newblock Prototypical cross-attention networks for multiple object tracking
  and segmentation.
\newblock {\em NeurIPS}, 2021.

\bibitem{ref-davis}
Anna Khoreva, Anna Rohrbach, and Bernt Schiele.
\newblock Video object segmentation with language referring expressions.
\newblock In {\em ACCV}, 2018.

\bibitem{PointRend}
Alexander Kirillov, Yuxin Wu, Kaiming He, and Ross Girshick.
\newblock Pointrend: Image segmentation as rendering.
\newblock In {\em CVPR}, 2020.

\bibitem{SiamRPNplusplus}
Bo Li, Wei Wu, Qiang Wang, Fangyi Zhang, Junliang Xing, and Junjie Yan.
\newblock {SiamRPN++}: {Evolution} of siamese visual tracking with very deep
  networks.
\newblock In {\em CVPR}, 2019.

\bibitem{SiameseRPN}
Bo Li, Junjie Yan, Wei Wu, Zheng Zhu, and Xiaolin Hu.
\newblock High performance visual tracking with siamese region proposal
  network.
\newblock In {\em CVPR}, 2018.

\bibitem{YOFO}
Dezhuang Li, Ruoqi Li, Lijun Wang, Yifan Wang, Jinqing Qi, Lu Zhang, Ting Liu,
  Qingquan Xu, and Huchuan Lu.
\newblock You only infer once: Cross-modal meta-transfer for referring video
  object segmentation.
\newblock In {\em AAAI}, 2022.

\bibitem{DN-DETR}
Feng Li, Hao Zhang, Shilong Liu, Jian Guo, Lionel~M Ni, and Lei Zhang.
\newblock Dn-detr: Accelerate detr training by introducing query denoising.
\newblock In {\em CVPR}, 2022.

\bibitem{GLIP}
Liunian~Harold Li, Pengchuan Zhang, Haotian Zhang, Jianwei Yang, Chunyuan Li,
  Yiwu Zhong, Lijuan Wang, Lu Yuan, Lei Zhang, Jenq-Neng Hwang, et~al.
\newblock Grounded language-image pre-training.
\newblock In {\em CVPR}, 2022.

\bibitem{RefTR}
Muchen Li and Leonid Sigal.
\newblock Referring transformer: A one-step approach to multi-task visual
  grounding.
\newblock {\em NeurIPS}, 2021.

\bibitem{CITD}
Chen Liang, Yu Wu, Tianfei Zhou, Wenguan Wang, Zongxin Yang, Yunchao Wei, and
  Yi Yang.
\newblock Rethinking cross-modal interaction from a top-down perspective for
  referring video object segmentation.
\newblock {\em arXiv preprint arXiv:2106.01061}, 2021.

\bibitem{RCCF}
Yue Liao, Si Liu, Guanbin Li, Fei Wang, Yanjie Chen, Chen Qian, and Bo Li.
\newblock A real-time cross-modality correlation filtering method for referring
  expression comprehension.
\newblock In {\em CVPR}, 2020.

\bibitem{ProposeReduce}
Huaijia Lin, Ruizheng Wu, Shu Liu, Jiangbo Lu, and Jiaya Jia.
\newblock Video instance segmentation with a propose-reduce paradigm.
\newblock In {\em ICCV}, 2021.

\bibitem{FPN}
Tsung-Yi Lin, Piotr Doll{\'a}r, Ross Girshick, Kaiming He, Bharath Hariharan,
  and Serge Belongie.
\newblock Feature pyramid networks for object detection.
\newblock In {\em CVPR}, 2017.

\bibitem{RetinaNet}
Tsung-Yi Lin, Priya Goyal, Ross Girshick, Kaiming He, and Piotr Doll{\'a}r.
\newblock Focal loss for dense object detection.
\newblock In {\em ICCV}, 2017.

\bibitem{COCO}
Tsung-Yi Lin, Michael Maire, Serge~J. Belongie, Lubomir~D. Bourdev, Ross~B.
  Girshick, James Hays, Pietro Perona, Deva Ramanan, Piotr Doll{\'a}r, and
  C.~Lawrence Zitnick.
\newblock {Microsoft COCO}: Common objects in context.
\newblock In {\em ECCV}, 2014.

\bibitem{NMTree}
Daqing Liu, Hanwang Zhang, Feng Wu, and Zheng-Jun Zha.
\newblock Learning to assemble neural module tree networks for visual
  grounding.
\newblock In {\em ICCV}, 2019.

\bibitem{CMPC+}
Si Liu, Tianrui Hui, Shaofei Huang, Yunchao Wei, Bo Li, and Guanbin Li.
\newblock Cross-modal progressive comprehension for referring segmentation.
\newblock {\em TPAMI}, 2021.

\bibitem{PANet}
Shu Liu, Lu Qi, Haifang Qin, Jianping Shi, and Jiaya Jia.
\newblock Path aggregation network for instance segmentation.
\newblock In {\em CVPR}, 2018.

\bibitem{CM-Att-Erase}
Xihui Liu, Zihao Wang, Jing Shao, Xiaogang Wang, and Hongsheng Li.
\newblock Improving referring expression grounding with cross-modal
  attention-guided erasing.
\newblock In {\em CVPR}, 2019.

\bibitem{ConvNeXt}
Zhuang Liu, Hanzi Mao, Chao-Yuan Wu, Christoph Feichtenhofer, Trevor Darrell,
  and Saining Xie.
\newblock {A} {ConvNet} for the 2020s.
\newblock In {\em CVPR}, 2022.

\bibitem{AdamW}
Ilya Loshchilov and Frank Hutter.
\newblock Decoupled weight decay regularization.
\newblock {\em arXiv preprint arXiv:1711.05101}, 2017.

\bibitem{Unified-IO}
Jiasen Lu, Christopher Clark, Rowan Zellers, Roozbeh Mottaghi, and Aniruddha
  Kembhavi.
\newblock {Unified-IO}: A unified model for vision, language, and multi-modal
  tasks.
\newblock {\em arXiv preprint arXiv:2206.08916}, 2022.

\bibitem{CGAN}
Gen Luo, Yiyi Zhou, Rongrong Ji, Xiaoshuai Sun, Jinsong Su, Chia-Wen Lin, and
  Qi Tian.
\newblock Cascade grouped attention network for referring expression
  segmentation.
\newblock In {\em ACMMM}, 2020.

\bibitem{MCN}
Gen Luo, Yiyi Zhou, Xiaoshuai Sun, Liujuan Cao, Chenglin Wu, Cheng Deng, and
  Rongrong Ji.
\newblock Multi-task collaborative network for joint referring expression
  comprehension and segmentation.
\newblock In {\em CVPR}, 2020.

\bibitem{RefCOCOg-g}
Junhua Mao, Jonathan Huang, Alexander Toshev, Oana Camburu, Alan~L Yuille, and
  Kevin Murphy.
\newblock Generation and comprehension of unambiguous object descriptions.
\newblock In {\em CVPR}, 2016.

\bibitem{ViTDet}
Yanghao Li~Hanzi Mao, Ross Girshick, and Kaiming He.
\newblock Exploring plain vision transformer backbones for object detection.
\newblock In {\em ECCV}, 2022.

\bibitem{keeptrack}
Christoph Mayer, Martin Danelljan, Danda~Pani Paudel, and Luc Van~Gool.
\newblock Learning target candidate association to keep track of what not to
  track.
\newblock In {\em ICCV}, 2021.

\bibitem{Trackformer}
Tim Meinhardt, Alexander Kirillov, Laura Leal-Taixe, and Christoph
  Feichtenhofer.
\newblock {Trackformer}: Multi-object tracking with transformers.
\newblock In {\em CVPR}, 2022.

\bibitem{MOT17}
Anton Milan, Laura Leal-Taix{\'e}, Ian Reid, Stefan Roth, and Konrad Schindler.
\newblock {MOT16}: A benchmark for multi-object tracking.
\newblock {\em arXiv preprint arXiv:1603.00831}, 2016.

\bibitem{DiceLoss}
Fausto Milletari, Nassir Navab, and Seyed-Ahmad Ahmadi.
\newblock V-net: Fully convolutional neural networks for volumetric medical
  image segmentation.
\newblock In {\em 3DV}, 2016.

\bibitem{trackingnet}
Matthias Muller, Adel Bibi, Silvio Giancola, Salman Alsubaihi, and Bernard
  Ghanem.
\newblock Trackingnet: A large-scale dataset and benchmark for object tracking
  in the wild.
\newblock In {\em ECCV}, 2018.

\bibitem{RefCOCOg-umd}
Varun~K Nagaraja, Vlad~I Morariu, and Larry~S Davis.
\newblock Modeling context between objects for referring expression
  understanding.
\newblock In {\em ECCV}, 2016.

\bibitem{STM}
Seoung~Wug Oh, Joon-Young Lee, Ning Xu, and Seon~Joo Kim.
\newblock Video object segmentation using space-time memory networks.
\newblock In {\em ICCV}, 2019.

\bibitem{QDTrack}
Jiangmiao Pang, Linlu Qiu, Xia Li, Haofeng Chen, Qi Li, Trevor Darrell, and
  Fisher Yu.
\newblock Quasi-dense similarity learning for multiple object tracking.
\newblock In {\em CVPR}, 2021.

\bibitem{DAVIS17}
Jordi Pont-Tuset, Federico Perazzi, Sergi Caelles, Pablo Arbel{\'a}ez, Alex
  Sorkine-Hornung, and Luc Van~Gool.
\newblock The 2017 {Davis} challenge on video object segmentation.
\newblock {\em arXiv preprint arXiv:1704.00675}, 2017.

\bibitem{OVIS}
Jiyang Qi, Yan Gao, Yao Hu, Xinggang Wang, Xiaoyu Liu, Xiang Bai, Serge
  Belongie, Alan Yuille, Philip~HS Torr, and Song Bai.
\newblock Occluded video instance segmentation: A benchmark.
\newblock {\em IJCV}, 2022.

\bibitem{YOLO}
Joseph Redmon, Santosh Divvala, Ross Girshick, and Ali Farhadi.
\newblock You only look once: Unified, real-time object detection.
\newblock In {\em CVPR}, 2015.

\bibitem{Gato}
Scott Reed, Konrad Zolna, Emilio Parisotto, Sergio~Gomez Colmenarejo, Alexander
  Novikov, Gabriel Barth-Maron, Mai Gimenez, Yury Sulsky, Jackie Kay,
  Jost~Tobias Springenberg, et~al.
\newblock A generalist agent.
\newblock {\em arXiv preprint arXiv:2205.06175}, 2022.

\bibitem{FasterRCNN}
Shaoqing Ren, Kaiming He, Ross Girshick, and Jian Sun.
\newblock {Faster R--CNN}: Towards real-time object detection with region
  proposal networks.
\newblock In {\em NeurIPS}, 2015.

\bibitem{GIoULoss}
Hamid Rezatofighi, Nathan Tsoi, JunYoung Gwak, Amir Sadeghian, Ian Reid, and
  Silvio Savarese.
\newblock Generalized intersection over union: A metric and a loss for bounding
  box regression.
\newblock In {\em CVPR}, 2019.

\bibitem{FRTM}
Andreas Robinson, Felix~Jaremo Lawin, Martin Danelljan, Fahad~Shahbaz Khan, and
  Michael Felsberg.
\newblock Learning fast and robust target models for video object segmentation.
\newblock In {\em CVPR}, 2020.

\bibitem{URVOS}
Seonguk Seo, Joon-Young Lee, and Bohyung Han.
\newblock {URVOS}: Unified referring video object segmentation network with a
  large-scale benchmark.
\newblock In {\em ECCV}, 2020.

\bibitem{INTERN}
Jing Shao, Siyu Chen, Yangguang Li, Kun Wang, Zhenfei Yin, Yinan He, Jianing
  Teng, Qinghong Sun, Mengya Gao, Jihao Liu, et~al.
\newblock {INTERN}: A new learning paradigm towards general vision.
\newblock {\em arXiv preprint arXiv:2111.08687}, 2021.

\bibitem{Objects365}
Shuai Shao, Zeming Li, Tianyuan Zhang, Chao Peng, Gang Yu, Xiangyu Zhang, Jing
  Li, and Jian Sun.
\newblock Objects365: A large-scale, high-quality dataset for object detection.
\newblock In {\em ICCV}, 2019.

\bibitem{sparsercnn}
Peize Sun, Rufeng Zhang, Yi Jiang, Tao Kong, Chenfeng Xu, Wei Zhan, Masayoshi
  Tomizuka, Lei Li, Zehuan Yuan, Changhu Wang, et~al.
\newblock {Sparse R-CNN}: End-to-end object detection with learnable proposals.
\newblock In {\em CVPR}, 2021.

\bibitem{CondInst}
Zhi Tian, Chunhua Shen, and Hao Chen.
\newblock Conditional convolutions for instance segmentation.
\newblock In {\em ECCV}, 2020.

\bibitem{FCOS}
Zhi Tian, Chunhua Shen, Hao Chen, and Tong He.
\newblock {FCOS}: Fully convolutional one-stage object detection.
\newblock In {\em ICCV}, 2019.

\bibitem{BoxInst}
Zhi Tian, Chunhua Shen, Xinlong Wang, and Hao Chen.
\newblock {BoxInst}: High-performance instance segmentation with box
  annotations.
\newblock In {\em CVPR}, 2021.

\bibitem{MOTS}
Paul Voigtlaender, Michael Krause, Aljosa Osep, Jonathon Luiten, Berin
  Balachandar~Gnana Sekar, Andreas Geiger, and Bastian Leibe.
\newblock {MOTS}: Multi-object tracking and segmentation.
\newblock In {\em CVPR}, 2019.

\bibitem{SiamRCNN}
Paul Voigtlaender, Jonathon Luiten, Philip~HS Torr, and Bastian Leibe.
\newblock Siam {R-CNN}: Visual tracking by re-detection.
\newblock In {\em CVPR}, 2020.

\bibitem{OFA-Ali}
Peng Wang, An Yang, Rui Men, Junyang Lin, Shuai Bai, Zhikang Li, Jianxin Ma,
  Chang Zhou, Jingren Zhou, and Hongxia Yang.
\newblock {OFA}: Unifying architectures, tasks, and modalities through a simple
  sequence-to-sequence learning framework.
\newblock In {\em ICML}, 2022.

\bibitem{SiamMask}
Qiang Wang, Li Zhang, Luca Bertinetto, Weiming Hu, and Philip H.~S. Torr.
\newblock Fast online object tracking and segmentation: {A} unifying approach.
\newblock In {\em CVPR}, 2019.

\bibitem{SOLO}
Xinlong Wang, Tao Kong, Chunhua Shen, Yuning Jiang, and Lei Li.
\newblock {SOLO}: Segmenting objects by locations.
\newblock In {\em ECCV}, 2020.

\bibitem{TNL-2K}
Xiao Wang, Xiujun Shu, Zhipeng Zhang, Bo Jiang, Yaowei Wang, Yonghong Tian, and
  Feng Wu.
\newblock Towards more flexible and accurate object tracking with natural
  language: Algorithms and benchmark.
\newblock In {\em CVPR}, 2021.

\bibitem{SOLOv2}
Xinlong Wang, Rufeng Zhang, Tao Kong, Lei Li, and Chunhua Shen.
\newblock {SOLOv2}: Dynamic and fast instance segmentation.
\newblock {\em NeurIPS}, 2020.

\bibitem{VISTR}
Yuqing Wang, Zhaoliang Xu, Xinlong Wang, Chunhua Shen, Baoshan Cheng, Hao Shen,
  and Huaxia Xia.
\newblock End-to-end video instance segmentation with transformers.
\newblock In {\em CVPR}, 2021.

\bibitem{JDE}
Zhongdao Wang, Liang Zheng, Yixuan Liu, Yali Li, and Shengjin Wang.
\newblock Towards real-time multi-object tracking.
\newblock In {\em ECCV}, 2020.

\bibitem{DeepSORT}
Nicolai Wojke, Alex Bewley, and Dietrich Paulus.
\newblock Simple online and realtime tracking with a deep association metric.
\newblock In {\em ICIP}, 2017.

\bibitem{SeqFormer}
J Wu, Y Jiang, S Bai, W Zhang, and X Bai.
\newblock Seqformer: Sequential transformer for video instance segmentation.
\newblock {\em ECCV}, 2021.

\bibitem{ReferFormer}
Jiannan Wu, Yi Jiang, Peize Sun, Zehuan Yuan, and Ping Luo.
\newblock Language as queries for referring video object segmentation.
\newblock In {\em CVPR}, 2022.

\bibitem{IDOL}
Junfeng Wu, Qihao Liu, Yi Jiang, Song Bai, Alan Yuille, and Xiang Bai.
\newblock In defense of online models for video instance segmentation.
\newblock {\em ECCV}, 2022.

\bibitem{OTB2015}
Yi Wu, Jongwoo Lim, and Ming~Hsuan Yang.
\newblock Object tracking benchmark.
\newblock {\em TPAMI}, 2015.

\bibitem{YoutubeVOS}
Ning Xu, Linjie Yang, Yuchen Fan, Dingcheng Yue, Yuchen Liang, Jianchao Yang,
  and Thomas Huang.
\newblock {YouTube-VOS}: A large-scale video object segmentation benchmark.
\newblock {\em arXiv preprint arXiv:1809.03327}, 2018.

\bibitem{PointTrackV2}
Zhenbo Xu, Wei Yang, Wei Zhang, Xiao Tan, Huan Huang, and Liusheng Huang.
\newblock Segment as points for efficient and effective online multi-object
  tracking and segmentation.
\newblock {\em TPAMI}, 2021.

\bibitem{Unicorn}
Bin Yan, Yi Jiang, Peize Sun, Dong Wang, Zehuan Yuan, Ping Luo, and Huchuan Lu.
\newblock Towards grand unification of object tracking.
\newblock In {\em ECCV}, 2022.

\bibitem{STARK}
Bin Yan, Houwen Peng, Jianlong Fu, Dong Wang, and Huchuan Lu.
\newblock Learning spatio-temporal transformer for visual tracking.
\newblock In {\em ICCV}, 2021.

\bibitem{AlphaRefine}
Bin Yan, Xinyu Zhang, Dong Wang, Huchuan Lu, and Xiaoyun Yang.
\newblock Alpha-refine: Boosting tracking performance by precise bounding box
  estimation.
\newblock In {\em CVPR}, 2021.

\bibitem{VIS}
Linjie Yang, Yuchen Fan, and Ning Xu.
\newblock Video instance segmentation.
\newblock In {\em ICCV}, 2019.

\bibitem{DGA}
Sibei Yang, Guanbin Li, and Yizhou Yu.
\newblock Dynamic graph attention for referring expression comprehension.
\newblock In {\em ICCV}, 2019.

\bibitem{RESC}
Zhengyuan Yang, Tianlang Chen, Liwei Wang, and Jiebo Luo.
\newblock Improving one-stage visual grounding by recursive sub-query
  construction.
\newblock In {\em ECCV}, 2020.

\bibitem{FAOA}
Zhengyuan Yang, Boqing Gong, Liwei Wang, Wenbing Huang, Dong Yu, and Jiebo Luo.
\newblock A fast and accurate one-stage approach to visual grounding.
\newblock In {\em ICCV}, 2019.

\bibitem{LAVT}
Zhao Yang, Jiaqi Wang, Yansong Tang, Kai Chen, Hengshuang Zhao, and Philip~HS
  Torr.
\newblock {LAVT}: Language-aware vision transformer for referring image
  segmentation.
\newblock In {\em CVPR}, 2022.

\bibitem{CFBI}
Zongxin Yang, Yunchao Wei, and Yi Yang.
\newblock Collaborative video object segmentation by foreground-background
  integration.
\newblock In {\em ECCV}, 2020.

\bibitem{OSTrack}
Botao Ye, Hong Chang, Bingpeng Ma, and Shiguang Shan.
\newblock Joint feature learning and relation modeling for tracking: A
  one-stream framework.
\newblock In {\em ECCV}, 2022.

\bibitem{CMSA}
Linwei Ye, Mrigank Rochan, Zhi Liu, and Yang Wang.
\newblock Cross-modal self-attention network for referring image segmentation.
\newblock In {\em CVPR}, 2019.

\bibitem{BDD100K}
Fisher Yu, Haofeng Chen, Xin Wang, Wenqi Xian, Yingying Chen, Fangchen Liu,
  Vashisht Madhavan, and Trevor Darrell.
\newblock {BDD100K}: A diverse driving dataset for heterogeneous multitask
  learning.
\newblock In {\em CVPR}, 2020.

\bibitem{RefCOCO&plus}
Licheng Yu, Patrick Poirson, Shan Yang, Alexander~C Berg, and Tamara~L Berg.
\newblock Modeling context in referring expressions.
\newblock In {\em ECCV}, 2016.

\bibitem{DINO}
Hao Zhang, Feng Li, Shilong Liu, Lei Zhang, Hang Su, Jun Zhu, Lionel~M Ni, and
  Heung-Yeung Shum.
\newblock {DINO}: Detr with improved denoising anchor boxes for end-to-end
  object detection.
\newblock {\em arXiv preprint arXiv:2203.03605}, 2022.

\bibitem{ATSS}
Shifeng Zhang, Cheng Chi, Yongqiang Yao, Zhen Lei, and Stan~Z Li.
\newblock Bridging the gap between anchor-based and anchor-free detection via
  adaptive training sample selection.
\newblock In {\em CVPR}, 2020.

\bibitem{bytetrack}
Yifu Zhang, Peize Sun, Yi Jiang, Dongdong Yu, Zehuan Yuan, Ping Luo, Wenyu Liu,
  and Xinggang Wang.
\newblock Bytetrack: Multi-object tracking by associating every detection box.
\newblock In {\em ECCV}, 2022.

\bibitem{FairMOT}
Yifu Zhang, Chunyu Wang, Xinggang Wang, Wenjun Zeng, and Wenyu Liu.
\newblock {FairMOT}: On the fairness of detection and re-identification in
  multiple object tracking.
\newblock {\em IJCV}, 2021.

\bibitem{TVOS}
Yizhuo Zhang, Zhirong Wu, Houwen Peng, and Stephen Lin.
\newblock A transductive approach for video object segmentation.
\newblock In {\em CVPR}, 2020.

\bibitem{Detic}
Xingyi Zhou, Rohit Girdhar, Armand Joulin, Philipp Kr{\"a}henb{\"u}hl, and
  Ishan Misra.
\newblock Detecting twenty-thousand classes using image-level supervision.
\newblock In {\em ECCV}, 2022.

\bibitem{CenterTrack}
Xingyi Zhou, Vladlen Koltun, and Philipp Kr{\"a}henb{\"u}hl.
\newblock Tracking objects as points.
\newblock In {\em ECCV}, 2020.

\bibitem{TRAR}
Yiyi Zhou, Tianhe Ren, Chaoyang Zhu, Xiaoshuai Sun, Jianzhuang Liu, Xinghao
  Ding, Mingliang Xu, and Rongrong Ji.
\newblock {TRAR}: Routing the attention spans in transformer for visual
  question answering.
\newblock In {\em ICCV}, 2021.

\bibitem{SeqTR}
Chaoyang Zhu, Yiyi Zhou, Yunhang Shen, Gen Luo, Xingjia Pan, Mingbao Lin, Chao
  Chen, Liujuan Cao, Xiaoshuai Sun, and Rongrong Ji.
\newblock {SeqTR}: A simple yet universal network for visual grounding.
\newblock {\em ECCV}, 2022.

\bibitem{DeformableDETR}
Xizhou Zhu, Weijie Su, Lewei Lu, Bin Li, Xiaogang Wang, and Jifeng Dai.
\newblock Deformable detr: Deformable transformers for end-to-end object
  detection.
\newblock In {\em ICLR}, 2020.

\bibitem{Uni-Perceiver}
Xizhou Zhu, Jinguo Zhu, Hao Li, Xiaoshi Wu, Hongsheng Li, Xiaohua Wang, and
  Jifeng Dai.
\newblock Uni-perceiver: Pre-training unified architecture for generic
  perception for zero-shot and few-shot tasks.
\newblock In {\em CVPR}, 2022.

\end{thebibliography}
}

\clearpage
\appendix

\section{Appendix}

\begin{table*}[t]
    \caption{Details in training. Step is the time to reduce the learning rate.}
    \centering
    {
\resizebox{1.0\linewidth}{!}{
\begin{tabular}{c|c|ccccc|cccc}

\toprule
Stage & Task & Dataset & Sampling Weight & Batch Size & Short & Long & Num GPU & Lr & Max Iter & Step \\

\midrule
\multirow{1}{*} {\uppercase\expandafter{\romannumeral1}}& \multirow{1}{*} {OD\&IS} & Objects365~\cite{Objects365} & 1 &  2 & $480\sim800$ & 1333 & 32 & 0.0002 & 340741 & 312346 \\
\midrule
\multirow{2}{*}{\uppercase\expandafter{\romannumeral2}} & OD\&IS & COCO~\cite{COCO} & 1 & 2 & $480\sim800$ & 1333 & \multirow{2}{*}{16} & \multirow{2}{*}{0.0002} & \multirow{2}{*}{91990} & \multirow{2}{*}{76658} \\
& REC\&RES & RefCOCO/g/+~\cite{RefCOCO&plus,RefCOCOg-umd} & 1 & 2 & $480\sim800$ & 1333 & & & & \\
\midrule
\multirow{14}{*}{\uppercase\expandafter{\romannumeral3}} & \multirow{5}{*}{SOT\&VOS} & LaSOT~\cite{LaSOT} & 0.20 & 2 & $480\sim800$ & 1333 & \multirow{14}{*}{16} & \multirow{14}{*}{0.0001} & \multirow{14}{*}{180000} & \multirow{14}{*}{150000} \\
& & GOT10K~\cite{GOT10K} & 0.20 & 2 & $480\sim800$ & 1333 & &&&\\
& & TrackingNet~\cite{trackingnet} & 0.20 & 2 & $480\sim800$ & 1333 & &&&\\
& & Youtube-VOS~\cite{YoutubeVOS} & 0.20 & 2 & $320\sim640$ & 768 &&&&\\
& & COCO~\cite{COCO} & 0.20 & 2 & $480\sim800$ & 1333 & &&&\\
\cline{2-7}
& \multirow{4}{*}{MOT\&MOTS} & BDD-obj-det~\cite{BDD100K} & 0.18 & 2 & $480\sim800$ & 1333  &&&&\\
& & BDD-box-track~\cite{BDD100K} & 0.72 & 2 & $480\sim800$ & 1333  &&&&\\
& & BDD-inst-seg~\cite{BDD100K} & 0.02 & 2 & $480\sim800$ & 1333  &&&&\\
& & BDD-seg-track~\cite{BDD100K} & 0.08 & 2 & $480\sim800$ & 1333  &&&&\\
\cline{2-7}
& \multirow{3}{*}{VIS} & Youtube-VIS-19~\cite{VIS} & 0.34 & 4 & $320\sim640$ & 768  &&&&\\
& & OVIS~\cite{OVIS} & 0.17 & 2 & $480\sim800$ & 1333  &&&&\\
& & COCO~\cite{COCO} & 0.51 & 2 & $480\sim800$ & 1333  &&&&\\
\cline{2-7}
& \multirow{2}{*}{R-VOS} & Ref-Youtube-VOS~\cite{URVOS} & 0.33 & 2 & $320\sim640$ & 768  &&&&\\
& & RefCOCO/g/+~\cite{RefCOCO&plus,RefCOCOg-umd} & 0.67 & 2 & $480\sim800$ & 1333  &&&&\\
\bottomrule
\end{tabular}}
    
}
    \label{tab:detail}
\end{table*}

In this appendix, we present more details about the training process and loss functions in ~\ref{sec-train-detail} and ~\ref{loss-func}, network architecture in ~\ref{sec-network}, as well as more analysis and visualizations for better understanding in ~\ref{sec-qualitative}.

\subsection{Training Process}
\label{sec-train-detail}
The detailed hyperparameters during training are shown in Tab~\ref{tab:detail}. The whole training process consists of three stages. In each stage, the \texttt{StepLR} learning rate scheduler is adopted. The learning rate drops by a factor of 10 after the given steps. For multi-dataset training, we follow the implementation of Detic~\cite{Detic}, which randomly samples data from different tasks and then computes them on different GPUs in one iteration. Besides, the multi-scale training technique is used across all datasets in all stages. Take the pre-training on Objects365~\cite{Objects365} as an example, the original images are resized such that the shortest side is at least 480 and at most 800 pixels while the longest side is at most 1333. We use this as the default setting except on Youtube-VOS~\cite{YoutubeVOS}, Youtube-VIS-2019~\cite{VIS}, and Ref-Youtube-VOS~\cite{URVOS}. A lower resolution with the shortest side ranging from 320 to 640 and the longest side not exceeding 768 is applied to these datasets~\cite{YoutubeVOS,VIS,URVOS}, following previous works~\cite{IDOL,ReferFormer,STCN}.

Specifically, in the first stage, the model is pretrained on Objects365~\cite{Objects365} for about 340K iterations (12 epochs) and the learning rate drops on the 11th epoch. In the second stage, we finetune UNINEXT on COCO~\cite{COCO} and RefCOCO/g/+~\cite{RefCOCO&plus,RefCOCOg-umd} jointly for 12 epochs. In the third stage, UNINEXT is further finetuned for diverse video-level tasks. To guarantee balanced performance on various benchmarks, we set the data sampling ratios as (SOT\&VOS):(MOT\&MOTS):VIS:R-VOS = 1:1:1:1. For each task, 45K iterations are allocated, thus bringing 180K iterations in total for the third stage. Besides, to avoid forgetting previously learned knowledge on image-level tasks, we also generate pseudo videos from COCO~\cite{COCO} and RefCOCO/g/+~\cite{RefCOCO&plus, RefCOCOg-umd} and mix them with training data of VIS~\cite{VIS,OVIS} and R-VOS~\cite{URVOS} respectively.

\subsection{Loss Functions}
\label{loss-func}
We present detailed loss functions described in Sec.~\ref{sub-sec-train-infer} for better readability. First, $\mathcal{L}_\mathrm{retrieve}$ and $\mathcal{L}_\mathrm{box}$ are used across all three stages. Second, to learn mask representations from coarse boxes~\cite{Objects365} and fine mask annotations~\cite{COCO,RefCOCO&plus,YoutubeVOS,VIS,URVOS}, UNINEXT uses $\mathcal{L}^\mathrm{boxinst}_\mathrm{mask}$ in the first stage and $\mathcal{L}_\mathrm{mask}$ in the next two stages respectively. Finally, to associate instances on different frames~\cite{BDD100K,VIS,OVIS}, UNINEXT additionally adopts $\mathcal{L}_\mathrm{embed}$ in the last stage.

\underline{\bm{$\mathcal{L}_\mathrm{retrieve}$}}. Given the raw instance-prompt matching score $s$, the normalized matching probability $p$ is computed as $p=\sigma(s)$, where $\sigma$ is sigmoid function. Then $\mathcal{L}_\mathrm{retrieve}$ can be written as the form of Focal loss~\cite{RetinaNet}.
 \eqnnm{flalpha}{\mathcal{L}_\mathrm{retrieve}(\pt) = - \at (1 - \pt)^\gamma \log (\pt).}
\eqnnm{pt}{\pt=\begin{cases} p &\text{if matched}\\ 1 - p &\text{otherwise.}\end{cases}}
$\gamma$ and $\alpha$ are 2 and 0.25 respectively.

\underline{\bm{$\mathcal{L}_\mathrm{box}$}}. Following DETR-like methods~\cite{DETR,DeformableDETR}, $\mathcal{L}_\mathrm{box}$ consists of two terms, GIoU Loss~\cite{GIoULoss} and $\ell_1$ loss:
 \eqnnm{loss_box}{\mathcal{L}_\mathrm{box}(b,\hat{b}) = \lambda_{giou}\mathcal{L}_\mathrm{giou}(b,\hat{b})+\lambda_{L_1}\Vert b-\hat{b} \Vert.}
\eqnnm{loss_giou}{\mathcal{L}_\mathrm{giou}(b,\hat{b})=1-IoU(b,\hat{b})+\frac{A^c(b,\hat{b})-U(b,\hat{b})}{A^c(b,\hat{b})},}
where $A^c(b,\hat{b})$ is the area of the smallest box containing $b$ and $\hat{b}$. $U(b,\hat{b})$ is the area of the union of $b$ and $\hat{b}$.

\underline{\bm{$\mathcal{L}_\mathrm{mask}$}}. For datasets with mask annotations~\cite{COCO,RefCOCO&plus,YoutubeVOS,VIS,URVOS}, Focal Loss~\cite{RetinaNet} and Dice Loss~\cite{DiceLoss} are adopted. 
\eqnnm{loss_mask}{\mathcal{L}_\mathrm{mask}(m,\hat{m}) = \lambda_{focal}\mathcal{L}_\mathrm{focal}(m,\hat{m})+\lambda_{dice}\mathcal{L}_\mathrm{dice}(m,\hat{m}).}
\eqnnm{loss_dice}{\mathcal{L}_\mathrm{dice}(m,\hat{m})=1-\frac{2m\hat{m}+1}{\hat{m}+m+1},}
where $m$ and $\hat{m}$ are binary GT masks and predicted masks after sigmoid activation respectively.

\underline{\bm{$\mathcal{L}^\mathrm{boxinst}_\mathrm{mask}$}}. For Objects365~\cite{Objects365} without mask annotations, UNINEXT uses Projection Loss and Pairwise Affinity Loss like BoxInst~\cite{BoxInst}, which can learn mask prediction only based on box-level annotations.
\eqnnm{loss_boxinst}{\mathcal{L}^\mathrm{boxinst}_\mathrm{mask}(b,\hat{m}) = \mathcal{L}_\mathrm{proj}(b,\hat{m})+\mathcal{L}_\mathrm{pairwise}(b,\hat{m}).}
\eqnnm{loss_proj}{
\begin{split}
\mathcal{L}_\mathrm{proj}(b,\hat{m})=&\mathcal{L}_\mathrm{dice}(\mathrm{proj_x}(b),\mathrm{proj_x}(\hat{m}))+\\
&\mathcal{L}_\mathrm{dice}(\mathrm{proj_y}(b),\mathrm{proj_y}(\hat{m})).
\end{split}
}
\eqnnm{loss_pairwise}{\mathcal{L}_\mathrm{pairwise} = -\frac{1}{N}\sum_{e \in E_{in}}\mathbbm{1}_{\{S_e \geq \tau\}}\log P(y_e = 1).}
\eqnnm{y_e}{
     P(y_e = 1) = \hat{m}_{i, j}
     \cdot
     \hat{m}_{k, l} + (1 - \hat{m}_{i, j})
     \cdot
     (1 - \hat{m}_{k, l}).}
\eqnnm{S_e}{S_e = S(c_{i, j}, c_{l, k}) = \exp\left(-\frac{||c_{i, j} - c_{l, k}||}{\theta}\right),}
where $y_e=1$ means the two pixels have the same ground-truth label. $S_e$ is the color similarity of the edge e. $c_{i,j}$ and $c_{l,k}$ are respectively the LAB color vectors of the two pixels $(i, j)$ and $(l, k)$ linked by the edge. $\theta$ is 2 in this work.

\underline{\bm{$\mathcal{L}_\mathrm{embed}$}}. UNINEXT uses contrastive loss~\cite{IDOL} to train discriminative embeddings for associating instances on different frames.
\eqnnm{loss_embed}{\mathcal{L}_\mathrm{embed} = \log [1+\sum_{\mathbf{k}^+}\sum_{\mathbf{k}^-}\exp(\mathbf{v} \cdot \mathbf{k}^-  - \mathbf{v} \cdot \mathbf{k}^+) ],}
where $k^+$ and $k^-$ are positive and negative feature embeddings from the reference frame. For each instance in the key frame, $v$ is the feature embedding with the lowest cost.

\begin{figure}[!t]
  \begin{center}
\includegraphics[width=1.0\linewidth]{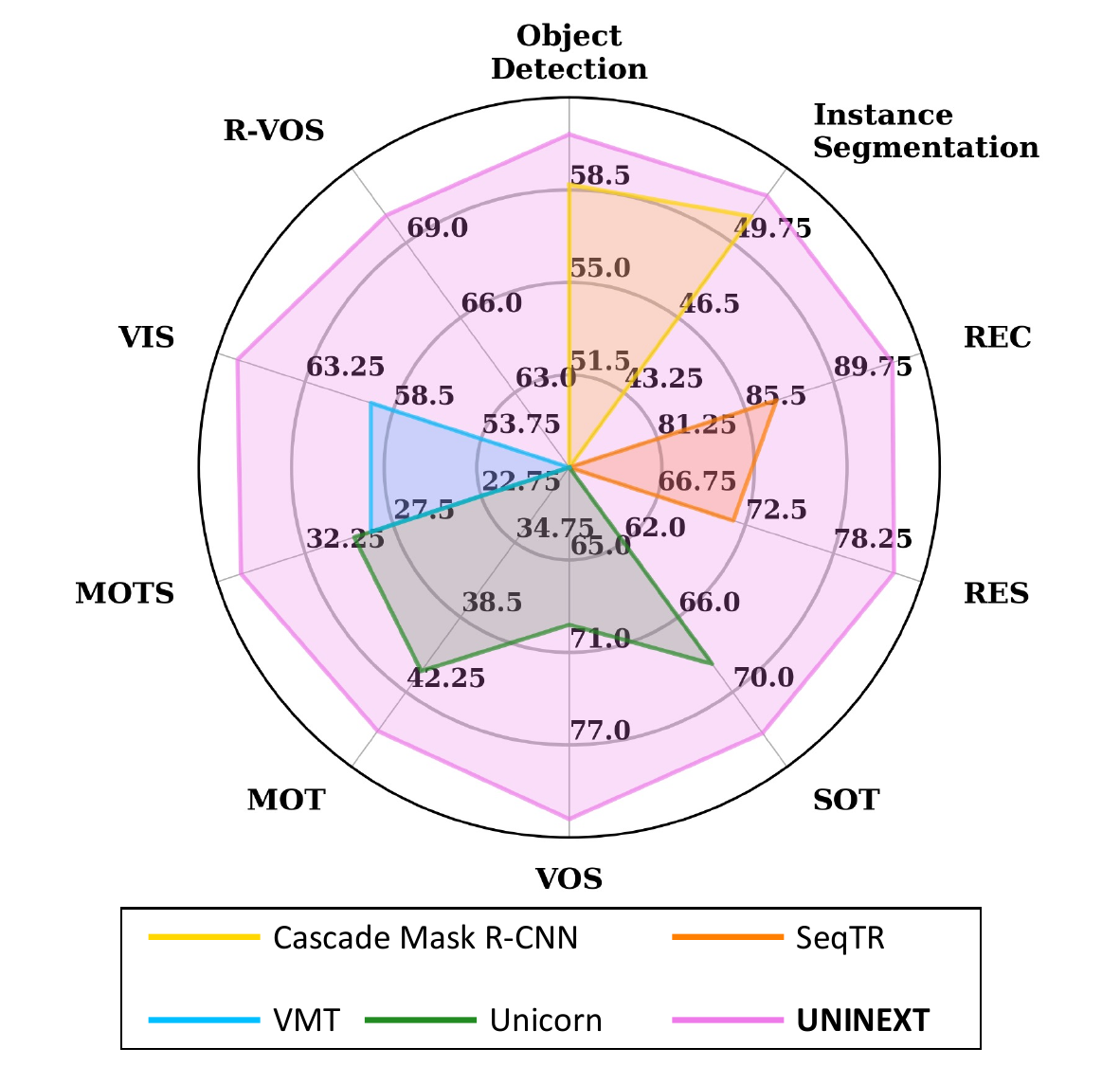}
  \end{center}
  \vspace{-5mm}
  \caption{Better view in color on screen.} 
  \label{fig-radar}
\vspace{-3mm}
\end{figure}

\begin{figure*}[!t]
  \begin{center}
\includegraphics[width=1.0\linewidth]{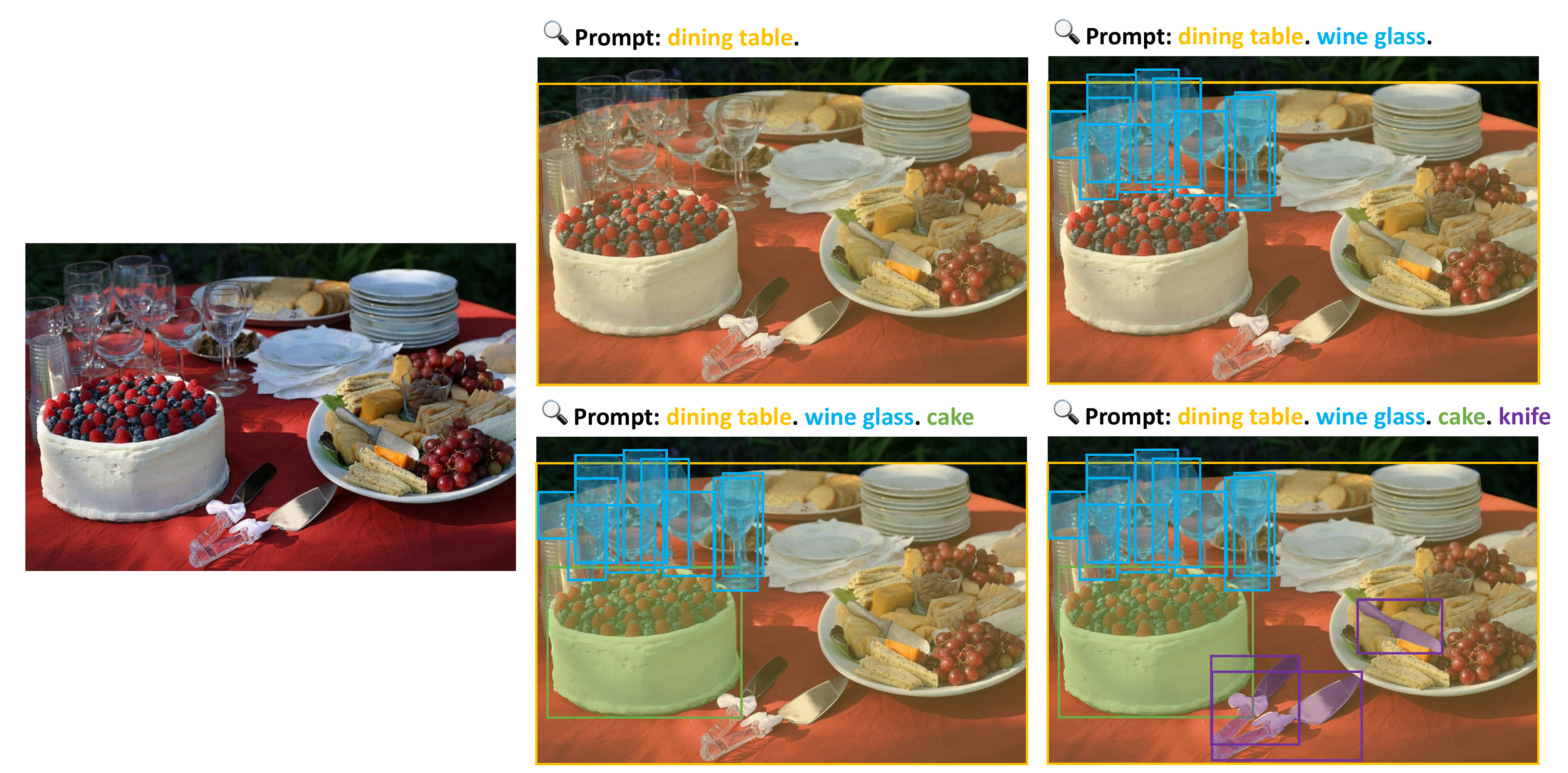}
  \end{center}
  \vspace{-5mm}
  \caption{Illustration of \textbf{retrieval by category names}. UNINEXT can flexibly perceive objects of different categories by changing the input prompts. Better view in color on screen.} 
  \label{fig-det-visual}
\vspace{-3mm}
\end{figure*}

\subsection{Network Architecture}
\label{sec-network}
To transform the enhanced visual features $F^{\prime}_v$ and prompt features $F^{\prime}_p$ into the final instance predictions, an encoder-decoder Transformer architecture is adopted. Based on the original architecture in two-stage Deformable DETR~\cite{DeformableDETR}, UNINEXT makes the following improvements:
\begin{itemize}
\item{\textbf{Introducing a mask head for segmentation.}} To predict high-quality masks, UNINEXT introduces a mask head~\cite{CondInst} based on dynamic convolutions. Specifically, first an MLP is used to transform instance embeddings into a group of parameters $\omega$. Then these parameters are used to perform three-layer $1\times1$ convolutions with feature maps, obtaining masks of instances.
\item{\textbf{Replacing one-to-one Hungarian matching with one-to-many SimOTA~\cite{YOLOX}.}} Traditional Hungarian matching forces one GT to be only assigned to one query, leaving most of the queries negative. UNINEXT uses SimOTA~\cite{YOLOX}, which enables multiple queries to be matched with one GT. This strategy can provide more positive samples and speed up convergence. During inference, UNINEXT uses NMS to remove duplicated predictions. 
\item{\textbf{Adding an IoU branch.}} UNINEXT adds an IoU branch to reflect the quality of the predicted boxes. During training, IoU does not affect the label assignment. During inference, the final scores are the geometric mean of the instance-prompt matching scores (after sigmoid) and the IoU scores.
\item{\textbf{Adding some techniques in DINO~\cite{DINO}.}} To further improve the performance, UNINEXT introduces some techniques~\cite{DINO}, including contrastive DN, mixed query selection, and look forward twice. 
\end{itemize}

\subsection{Analysis and Visualizations}
\label{sec-qualitative}

\textbf{Analysis}. We compare UNINEXT with other competitive counterparts, which can handle multiple instance-level perception tasks. The opponents include Cascade Mask R-CNN~\cite{CascadeRCNN} for object detection and instance segmentation, SeqTR~\cite{SeqTR} for REC and RES, VMT~\cite{VMT} for MOTS and VIS, and Unicorn~\cite{Unicorn} for SOT, VOS, MOT, and MOTS. As shown in Figure~\ref{fig-radar}, UNINEXT outperforms them and achieve state-of-the-art performance on all 10 tasks. 

\textbf{Retrieval by Category Names}. As shown in Figure~\ref{fig-det-visual}, UNINEXT can flexibly detect and segment objects of different categories by taking the corresponding category names as the prompts. For example, when taking ``dining table. wine glass. cake. knife'' as the prompts, UNINEXT would only perceive dining tables, wine glasses, cakes, and knives. Furthermore, benefiting from the flexible retrieval formulation, UNINEXT also has the potential for zero-shot (open-vocabulary) object detection. However, open-vocabulary object detection is beyond the scope of our paper and we leave it for future works.

\begin{figure*}[!t]
  \begin{center}
\includegraphics[width=1.0\linewidth]{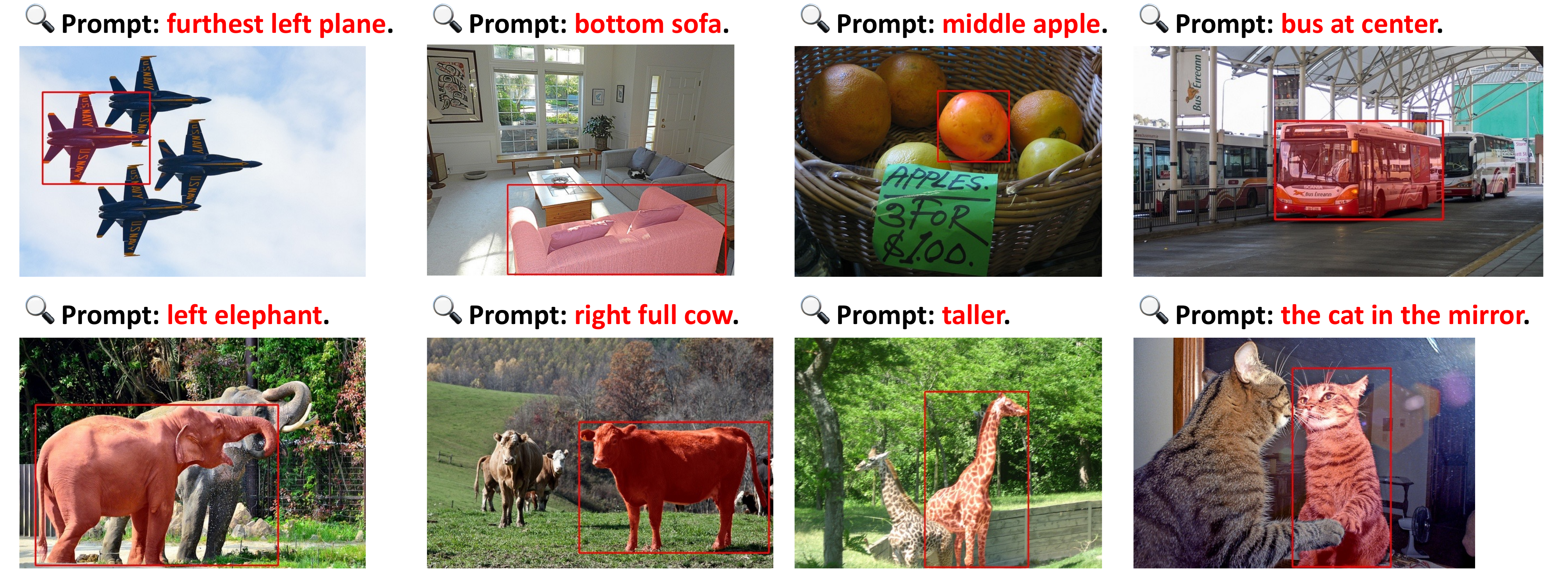}
  \end{center}
  \vspace{-5mm}
  \caption{Illustration of \textbf{retrieval by language expressions}. Better view in color on screen.}
  \label{fig-rec-res}
\vspace{-3mm}
\end{figure*}

\textbf{Retrieval by Language Expressions}. We provide some visualizations for retrieval by language expressions in Figure~\ref{fig-rec-res}. UNINEXT can accurately locate the target referred by the given language expression when there are many similar distractors. This demonstrates that our method can not only perceive objects but also understand their relationships in positions (left, middle, right, etc) and sizes (taller, etc). 

\begin{figure*}[!t]
  \begin{center}
\includegraphics[width=0.9\linewidth]{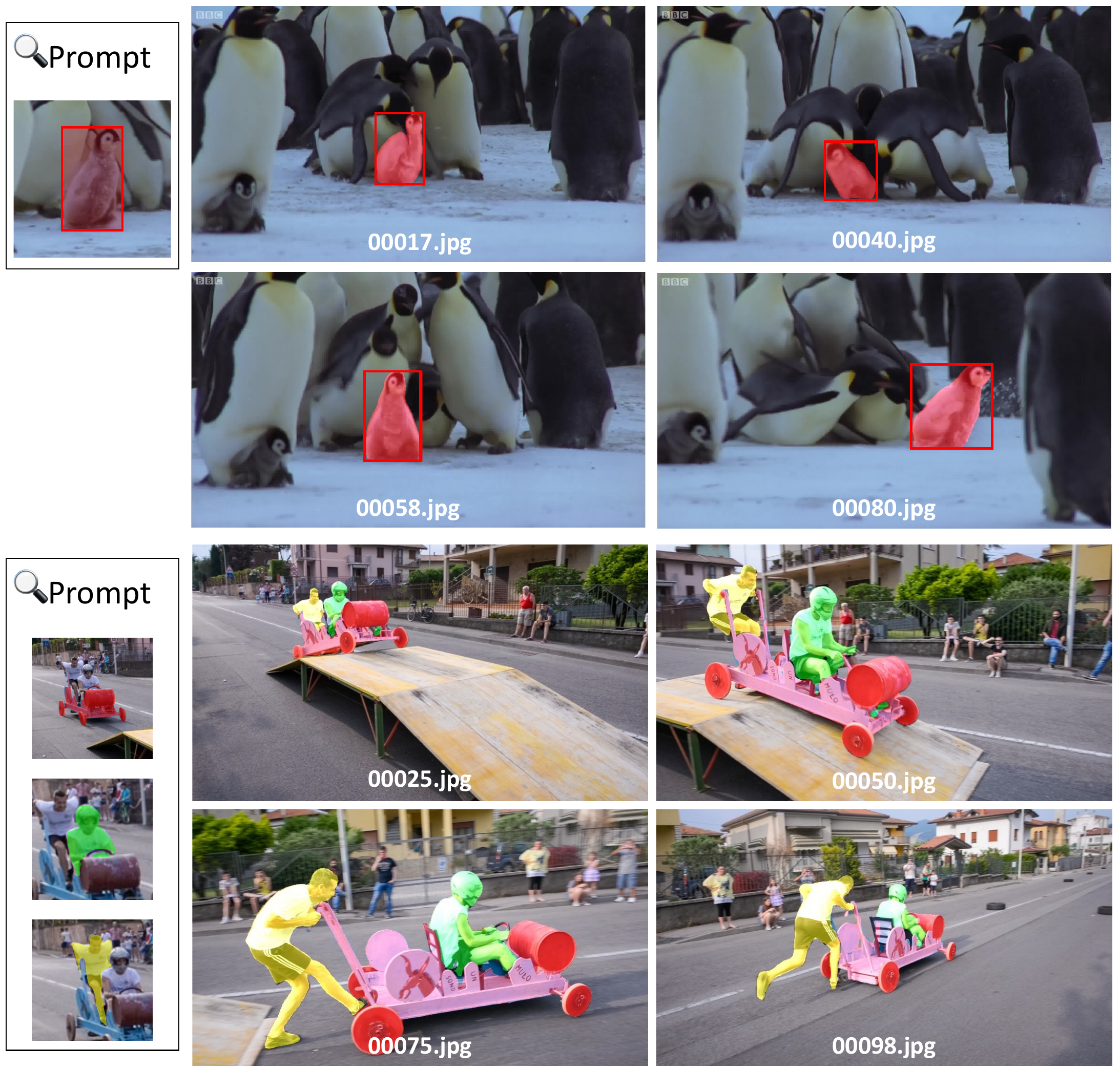}
  \end{center}
  \vspace{-5mm}
  \caption{Illustration of \textbf{retrieval by target annotations}. UNINEXT can flexibly perceive different objects according to the box or mask annotations given in the first frame. Better view in color on screen.} 
  \label{fig-vos}
\vspace{-3mm}
\end{figure*}

\textbf{Retrieval by Target Annotations}. Our method supports annotations in formats of both boxes (SOT) and masks (VOS). Although there is only box-level annotation for SOT, we obtain the target prior by filling the region within the given box with $1$ and leaving other regions as $0$. As shown in Figure~\ref{fig-vos}, UNINEXT can precisely track and segment the targets in complex scenarios, given the annotation in the first frame.

\end{document}